\documentclass{article}

\usepackage[english]{babel}

\usepackage[letterpaper,top=2cm,bottom=2cm,left=3cm,right=3cm,marginparwidth=1.75cm]{geometry}

\usepackage{amsmath}
\usepackage{graphicx}
\usepackage{longtable}
\usepackage{booktabs}
\usepackage{array}
\usepackage{caption}
\usepackage[colorlinks=true, allcolors=blue]{hyperref}

\title{Large-Scale AI and Foundation Models for Neuroscience: A Comprehensive Review}

\author{%
\begin{minipage}{\textwidth}
\centering
\footnotesize
Shihao Yang\textsuperscript{1},
Xiying Huang\textsuperscript{1},
Danilo Bernardo\textsuperscript{2},
Jun\mbox{-}En Ding\textsuperscript{1},\\
Andrew Michael\textsuperscript{3},
Guoan Wang\textsuperscript{1},\\
Jingmei Yang\textsuperscript{4},
Alison Anderson\textsuperscript{5},
Dinesh Giritharan\textsuperscript{5,6},\\
Patrick Kwan\textsuperscript{5,6},
Ashish Raj\textsuperscript{7},
Yu Zhang\textsuperscript{8,9,10},
Feng Liu\textsuperscript{1}\thanks{Corresponding author: fliu22@stevens.edu.}
\\[4pt]
\textsuperscript{1}Department of Systems Engineering, Stevens Institute of Technology, Hoboken, NJ, USA\\
\textsuperscript{2}Department of Neurology and Weill Institute for Neurosciences, University of California San Francisco, \\
San Francisco, CA, USA\\
\textsuperscript{3}Duke Institute for Brain Sciences, Duke University, Durham, NC, USA\\
\textsuperscript{4}Division of Systems Engineering, Department of Electrical and Computer Engineering, Boston University, 
\\ 
Boston, MA, USA\\
\textsuperscript{5}Department of Neuroscience, School of Translational Medicine, Monash University, Melbourne, Victoria, Australia\\
\textsuperscript{6}Department of Neurology, Alfred Hospital, Melbourne, Victoria, Australia\\
\textsuperscript{7}Department of Radiology and Biomedical Imaging, University of California, San Francisco, CA, USA\\
\textsuperscript{8}Department of Psychiatry and Behavioral Sciences, School of Medicine, Stanford University, Stanford, CA, USA\\
\textsuperscript{9}Wu Tsai Neurosciences Institute, Stanford University, Stanford, CA, USA\\
\textsuperscript{10}Stanford Institute for Human-Centered AI, Stanford University, Stanford, CA, USA
\end{minipage}
}

\date{}

\begin{document}
\maketitle

\begin{abstract}
The development of large-scale artificial intelligence (AI) models is influencing neuroscience research by enabling end-to-end learning from raw brain signals and neural data. In this paper, we review applications of large-scale AI models across five major neuroscience domains: neuroimaging and data processing, brain-computer interfaces and neural decoding, clinical decision support and translational frameworks, and disease-specific applications across neurological and psychiatric disorders. These models show potential to address major computational neuroscience challenges, including multimodal neural data integration, spatiotemporal pattern interpretation, and the development of translational frameworks for clinical research. Moreover, the interaction between neuroscience and AI has become increasingly reciprocal, as biologically informed architectural constraints are now incorporated to develop more interpretable and computationally efficient models. This review highlights both the promise of such technologies and critical implementation considerations, with particular emphasis on rigorous evaluation frameworks, effective integration of domain knowledge, prospective clinical validation, and comprehensive ethical guidelines. Finally, a systematic listing of critical neuroscience datasets used to develop and evaluate large-scale AI models across diverse research applications is provided.
\end{abstract}

\section{Introduction}

Foundation models (FMs) and large language models (LLMs) represent a shift in how AI systems engage with neuroscience data, moving from hand-engineered features toward learned representations derived from large datasets~\cite{bommasani2021opportunities, wang2025large}. Classical computational neuroscience relied on domain-specific feature engineering followed by conventional machine learning methods, such as support vector machines, linear discriminant analysis, and ensemble methods~\cite{lotte2007review}. In contrast, transformer-based architectures and self-supervised pretraining enable end-to-end learning of hierarchical representations directly from raw neural recordings~\cite{schirrmeister2017deep}. Neuroimaging data in this context encompasses electrophysiological and structural or functional imaging modalities, including EEG, MEG, sEEG, ECoG, CT, and fMRI, each operating at different spatiotemporal resolutions and providing complementary perspectives on brain architecture and neural dynamics. This architectural evolution carries both promise and unresolved methodological tensions. Proponents argue that FMs generalize across experimental conditions, subjects, and recording modalities by capturing intricate spatiotemporal dependencies in neural data~\cite{wang2025foundation}. However, most validation occurs within controlled research settings on curated benchmarks, leaving cross-site robustness, real-world deployment challenges, and failure mode characterization largely unaddressed. 

Self-supervised learning (SSL) has emerged as a response to labeled data scarcity in neuroscience~\cite{he2020momentum}. Brain foundation models (BFMs) leverage contrastive learning and transformer architectures to learn representations from unlabeled neural recordings, with applications ranging from cross-subject decoding to neurological biomarker discovery~\cite{zhang2023brant, kostas2021bendr}. BENDR, for instance, integrates transformers with contrastive objectives to produce EEG representations that transfer across tasks~\cite{kostas2021bendr}. These methods are credited with enabling systems that decode intended speech from neural signals, predict seizure onset, and detect neurodegenerative biomarkers before clinical symptom onset~\cite{chen2025brain, willett2021high}. Yet the gap between reported research performance and clinically validated systems remains substantial. Speech decoding "at near-conversational speeds" refers to controlled laboratory settings with invasive intracranial recordings, not non-invasive deployable devices. Seizure prediction "with remarkable accuracy" often reflects retrospective analysis on single-site datasets without prospective multi-site validation. Biomarker claims for neurodegenerative diseases describe statistical associations in research cohorts, not diagnostic tools approved for clinical use. The distinction between technical feasibility in research contexts and clinical utility in deployment settings is critical but frequently elided.

Multimodal integration represents a major application area for large-scale AI in neuroscience. Vision-language models such as CLIP generate cross-modal semantic embeddings, enabling alignment of neural activity patterns with textual descriptions or visual stimuli~\cite{radford2021learning}. In clinical decision support, FMs integrate neuroimaging, genetic data, clinical histories, and physiological monitoring to inform treatment recommendations~\cite{pedersen2020artificial, wang2024mindbridge}. Generative models employing latent diffusion architectures reconstruct visual experiences from fMRI activity~\cite{ozcelik2023natural}, while diffusion-based data augmentation synthesizes EEG signals to address data scarcity~\cite{soingern2023data}. Cross-modal fusion architectures combine neuroimaging modalities, genomic information, clinical metadata, and behavioral measures to model brain function and pathology~\cite{simon2024future, lee2024multimodal}.

The interplay between neuroscience and AI extends beyond application to mutual methodological influence. Brain-inspired architectures incorporating biological constraints and neural coding principles have improved model interpretability, robustness, and computational efficiency in some contexts~\cite{richards2019deep, cui2024toward}. LLMs demonstrate capacity to predict neuroscience experiment outcomes, process scientific literature, and generate hypotheses~\cite{luo2025large}, though the practical impact of such capabilities on experimental neuroscience workflows remains to be established. Neuromorphic processors such as Intel's Loihi and IBM's TrueNorth implement brain-inspired computing paradigms for energy-efficient AI~\cite{merolla2014million, davies2018loihi}. Research on spiking neural networks, neuromorphic computing, and biologically plausible attention mechanisms aims to advance both understanding of biological intelligence and development of efficient AI systems~\cite{davies2018loihi, roy2019towards, zador2023catalyzing}. These contributions are significant but do not constitute a wholesale transformation of neuroscience methodology. Instead, they represent incremental advances within specific subdomains.

The trajectory from classical machine learning to contemporary multimodal and generative models has expanded the scope of neural decoding, brain function interpretation, and biologically inspired AI development~\cite{bzdok2024data, doerig2023neuroconnectionist, cichy2019deep, kietzmann2019recurrence}. These advances reflect genuine technical progress but also highlight persistent challenges. Multimodal alignment in controlled experimental paradigms, characterized by carefully standardized stimuli, acquisition protocols, and subject populations, does not guarantee robustness to the variability encountered in clinical practice. Clinical decision support systems that "integrate multimodal patient data" in research prototypes face regulatory, liability, and workflow integration barriers before clinical adoption. Visual reconstruction from fMRI achieves high fidelity on specific stimulus sets within controlled scanner environments, but it remains unclear whether these methods generalize to naturalistic viewing conditions, diverse scanner types, and heterogeneous patient populations. Synthetic data augmentation improves performance on held-out test sets drawn from the same distribution as training data but may not address fundamental data limitations such as site-specific biases, demographic imbalances, or pathological heterogeneity.

Existing surveys have largely summarized technical advances within specific neuroscience subdomains, but have not systematically examined how reported performance relates to validation scope, deployment readiness, and the strength of supporting evidence across models and applications. In this review, we use "large-scale AI models" to encompass FMs, LLMs, vision-language models, and other transformer-based and generative architectures trained with millions or billions of parameters. To ensure a rigorous analysis of these diverse architectures, we adopt an operational classification framework that distinguishes between three fundamental categories: General-Purpose Foundation Models, which aim for universal cross-modal representation; Modality-Specific Base Models, which optimize encoding for specific signals like EEG or fMRI; and Task-Oriented Large-Scale Models, which prioritize performance on specific clinical endpoints. This distinction allows for a more precise assessment of their respective generalization properties and deployment readiness. This review organizes applications into five domains in Section 3. First, neuroimaging and data processing, including brain structure analysis, electrophysiological signal interpretation, and multimodal integration. Second, brain-computer interfaces and neural decoding systems translate neural activity into actionable outputs. %Third, molecular neuroscience and genomic modeling, covering variant interpretation, genetic risk prediction, and disease mechanism discovery. 
Third, clinical support and translational frameworks, including decision support systems, knowledge-driven AI, and cognitive modeling. Fourth, applications targeting neurological and psychiatric disorders, including neurodegenerative conditions, acute injuries, brain tumors, neuropsychiatric illnesses, and neurodevelopmental disorders. For each domain, we analyze reported performance, validation breadth, and deployment readiness, highlighting both substantive methodological contributions and the gap between technical capabilities and the available evidence supporting clinical utility. Section 4 catalogs public datasets that enable large-scale model development and evaluation. Section 5 synthesizes recurring challenges, including cross-site generalization, multimodal integration, interpretability, and the gap between retrospective validation and prospective clinical deployment, concluding with priorities for responsible translation into validated tools and impactful research applications.

% \begin{figure}[!htbp]
%     \centering
%     \setlength{\fboxrule}{1pt}
%     \includegraphics[width=1\linewidth]{figures/intro.pdf}
%     \captionof{Neuroscience AI Landscape: Developmental Trajectory and Functional Framework.}
%     \label{fig:intro}
% \end{figure}

\begin{figure}
\centering
\includegraphics[width=0.95\linewidth]{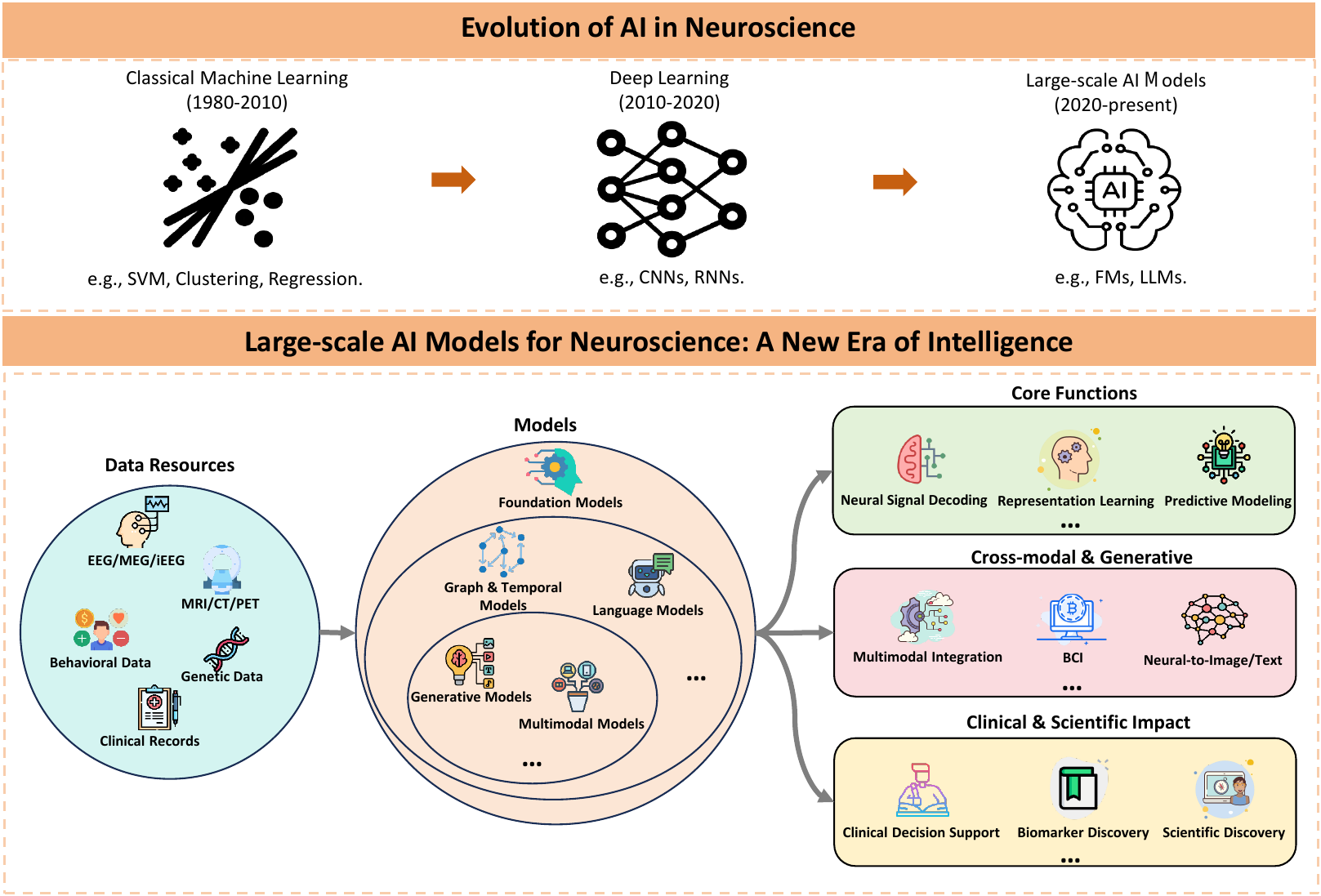}
\caption{The evolving landscape and functional framework of AI in neuroscience.
The upper panel summarizes the transition from classical machine learning to modern large-scale AI models. The lower panel illustrates how heterogeneous data resources are integrated within contemporary AI systems. Foundation models represent a general training paradigm, while language, graph-based, and temporal models reflect complementary architectural dimensions. These components may be combined within multimodal and generative systems to support core functions, cross-modal applications, and clinical and scientific use cases.}
\label{fig:intro}
\end{figure}

\section{Background and problem formulation}
\subsection{Theoretical foundations of large-scale AI models}

\subsubsection{From classical machine learning to FMs}
The trajectory from task-specific machine learning to FMs in neuroscience reflects broader shifts in computational capacity and data availability rather than fundamental algorithmic innovation. Early approaches relied on hand-engineered features from neurophysiological signals, such as support vector machines, linear discriminant analysis, and ensemble methods, achieving domain-specific success at the cost of generalization~\cite{lotte2007review}. Deep learning eliminated manual feature engineering through end-to-end hierarchical representation learning from raw neural recordings~\cite{schirrmeister2017deep}, yet remained constrained by annotation requirements and task specificity. FMs extended this paradigm by leveraging self-supervised pretraining on large unlabeled corpora, enabling transfer across tasks with minimal fine-tuning~\cite{bommasani2021opportunities}. However, this evolutionary narrative obscures persistent methodological tensions. An overview of neuroscience AI models from fundamental computational functions through cross-modal capabilities is contextualized in Fig.~\ref{fig:intro}, though such taxonomies risk conflating architectural diversity with validated clinical utility.

\subsubsection{Scaling laws and core principles}
Transformer architectures leverage self-attention mechanisms to capture long-range dependencies in neural data~\cite{vaswani2017attention}, a design principle particularly relevant for high-dimensional spatiotemporal signals. Empirical scaling laws suggest that model performance improves predictably with increases in parameters, training data, and compute~\cite{kaplan2020scaling}, leading to claims that larger models trained on more neural data will necessarily improve decoding accuracy and cross-subject transfer. This assumption merits critical examination. Scaling laws derived from language modeling may not transfer directly to neuroscience contexts where data scarcity, inter-subject variability, and non-stationarity introduce domain-specific bottlenecks. Emergent capabilities such as few-shot learning and zero-shot generalization across recording modalities remain largely aspirational rather than empirically validated in clinical neuroscience settings. The relationship between model scale and biological interpretability is inverse, that larger models often sacrifice mechanistic transparency for predictive accuracy, creating a fundamental tension between performance metrics and scientific understanding.

\subsection{Operational classification of neuroscience AI models}
To address the ambiguity between general representation learning and task-specific problem solving, we organize the models reviewed in this article into three operational categories based on their primary objectives and generalization scope (Table~\ref{tab:classification}).

\textbf{General-purpose foundation models} aim to learn broadly transferable neural representations from large, heterogeneous datasets spanning multiple modalities and tasks (e.g., BrainLM). Their central goal is to capture intrinsic neural dynamics that support zero-shot or few-shot adaptation to previously unseen tasks.

\textbf{Modality-specific base models} focus on learning high-fidelity representations for a single data modality, such as fMRI or EEG. By exploiting modality-specific physical constraints and spatiotemporal structure, these models often achieve robust within-modality generalization, but do not inherently transfer across modalities (e.g., EEGFormer).

\textbf{Task-oriented large-scale models} are designed to optimize performance for well-defined clinical or scientific endpoints (e.g., SETR-PKD). Although such systems may leverage large-scale pretrained backbones, their architectures and fine-tuning strategies are tightly coupled to specific applications, prioritizing task-level precision over broad transferability.

{\footnotesize
\setlength{\tabcolsep}{4pt}
\begin{longtable}{p{4.5cm} p{3.8cm} p{3.2cm} p{3.7cm}}
\caption{Operational classification framework for large-scale AI models in neuroscience} \label{tab:classification} \\
\toprule
\textbf{Model Category} & \textbf{Primary Objective} & \textbf{Generalization Properties} & \textbf{Representative Examples} \\
\midrule
\endfirsthead
\caption[]{(continued)} \\
\toprule
\textbf{Model Category} & \textbf{Primary Objective} & \textbf{Generalization Properties} & \textbf{Representative Examples} \\
\midrule
\endhead
\bottomrule
\multicolumn{4}{r}{Continued on next page} \\
\endfoot
\bottomrule
\endlastfoot
\textbf{General-Purpose Foundation Models} & Learn universal, transferable representations across diverse tasks and modalities & \textbf{High:} Capable of zero-shot or few-shot transfer to unseen tasks/datasets & BrainWave~\cite{yuan2024brainwave}, BrainLM~\cite{caro2023brainlm} \\
\textbf{Modality-Specific Base Models} & Encode high-fidelity features specific to a signal type (e.g., EEG dynamics) & \textbf{Medium:} Robust within the specific modality; limited cross-modal transfer & EEGFormer~\cite{chen2024eegformer}, BRANT~\cite{zhang2023brant} \\
\textbf{Task-Oriented Large-Scale Models} & Maximize predictive performance on specific clinical/scientific endpoints & \textbf{Low:} Specialized for defined tasks; limited robustness to distribution shifts & SETR-PKD~\cite{mehta2023privacy}, EpiSemoLLM~\cite{yang2024episemollm} \\
\end{longtable}
}

\subsection{Architectural frameworks and design principles}

\subsubsection{Transformer architecture in neuroscience applications}

Transformers have been adapted across multiple neuroscience domains, as detailed in Table~\ref{tab:transformer_applications}. In EEG and MEG analysis, models like BRANT apply multi-layer transformers to intracranial recordings for cross-patient neural decoding~\cite{zhang2023brant}, while CBraMod employs cross-attention structures for EEG-to-cognitive-state decoding~\cite{wang2024cbramod}. For MRI, systems such as MindBridge and MindEye2 align voxel patterns with visual embeddings through transformer layers, enabling image reconstruction from fMRI~\cite{wang2024mindbridge, scotti2024mindeye2}. Clinical NLP applications leverage pretrained language models for entity linking and reasoning in medical contexts~\cite{cui2023neuro, hopkins2024atlasgpt}.

These applications demonstrate architectural versatility but raise methodological concerns. Most transformer-based neuroscience models report performance on single-site datasets with limited external validation. BRANT's generalization claims across "patients and tasks" require qualification: the study evaluates on intracranial EEG from epilepsy patients within a single clinical protocol, not across diverse institutions or recording systems. MindEye2's "minimal training data" framing obscures the substantial pretraining requirements and reliance on shared-subject paradigms that may not extend to naturalistic settings. The attention mechanism's purported advantage in "handling brain shape variability and imaging resolution differences" conflates model flexibility with validated robustness, few studies systematically test performance degradation under site-specific acquisition protocols or scanner heterogeneity.

{\footnotesize
\setlength{\tabcolsep}{4pt}
\begin{longtable}{p{4cm} p{4cm} p{4cm} p{3cm}}
    \caption{Transformer architecture applications across neuroscience domains} \label{tab:transformer_applications} \\
    \toprule
    \textbf{Domain} & \textbf{Key Challenges} & \textbf{Transformer Advantages} & \textbf{Model Examples} \\
    \midrule
    \endfirsthead
    \caption[]{(continued)} \\
    \toprule
    \textbf{Domain} & \textbf{Key Challenges} & \textbf{Transformer Advantages} & \textbf{Representative Models} \\
    \midrule
    \endhead
    \bottomrule
    \multicolumn{4}{r}{Continued on next page} \\
    \endfoot
    \bottomrule
    \endlastfoot
\textbf{EEG/MEG Analysis} & Non-stationarity, temporal dynamics, artifacts & Long-range temporal modeling, self-attention for time-series & BRANT~\cite{zhang2023brant}, CBraMod~\cite{wang2024cbramod}, BENDR~\cite{kostas2021bendr} \\
\textbf{MRI Decoding} & High dimensionality, inter-subject variability & Spatiotemporal attention, contextual encoding & MindBridge~\cite{wang2024mindbridge}, MindEye2~\cite{scotti2024mindeye2} \\
\textbf{Structural MRI} & Anatomical variability, low contrast structures & Patch tokenization, cross-slice attention & BrainSegFounder~\cite{cox2024brainsegfounder}, AnatCL~\cite{barbano2024anatomical} \\
\textbf{Multimodal Neuroimaging} & Cross-modal alignment, data heterogeneity & Cross-modal attention, joint representation learning & BrainCLIP~\cite{ma2025brainclip}, MultiViT~\cite{bi2024multimodal} \\
\textbf{Clinical NLP} & Medical terminology, reasoning, sparse context & Pretrained LLMs, in-context learning, entity linking & NeuroGPT~\cite{cui2023neuro}, AtlasGPT~\cite{hopkins2024atlasgpt} \\
\textbf{Genomics} & Predicting function of genetic variants & Capturing long-range genomic interactions & AlphaGenome~\cite{avsec2025alphagenome}, Enformer~\cite{avsec2021effective} \\
\textbf{Seizure Video Analysis} & Variable semiology, temporal dynamics, environmental variability & Long-range temporal modelling, robust feature representation, multimodal integration & SETR-PKD~\cite{mehta2023privacy}, VSViG~\cite{xu2024vsvig} \\
\end{longtable}
}

\subsubsection{Multimodal integration architectures}
Multimodal architectures in neuroscience aim to integrate heterogeneous data types, such as imaging, electrophysiology, genomics, behavior, into joint representations that exceed single-modality capabilities~\cite{radford2021learning}. Table~\ref{tab:multimodal_fusion} categorizes fusion strategies by integration level. Early fusion concatenates raw modality tokens before encoding; mid-level fusion employs shared attention layers across modalities; late fusion combines separately encoded representations at the decision stage; prompt-based approaches use text to guide cross-modal attention; contrastive alignment learns shared embedding spaces; hierarchical fusion stages modalities sequentially.

The fundamental challenge in multimodal fusion lies not in architectural choice but in establishing whether integration genuinely improves clinical outcomes versus merely improving proxy metrics on research benchmarks. BrainCLIP aligns fMRI voxel patterns with visual embeddings through contrastive learning~\cite{ma2025brainclip}, demonstrating semantic alignment on curated datasets. Whether such alignment transfers to diagnostic contexts with naturalistic stimuli and uncontrolled acquisition conditions remains unvalidated. MindBridge claims "significant improvements in reconstruction fidelity and generalization capabilities"~\cite{wang2024mindbridge} based on within-study cross-validation, yet cross-site external validation that testing on data from institutions not represented in training is absent. This pattern recurs, where models report improvements over baselines in controlled experimental paradigms but provide limited evidence of robustness to the site-specific biases, protocol variations, and population heterogeneity that characterize clinical deployment.

{\footnotesize
\setlength{\tabcolsep}{4pt}
\begin{longtable}{p{3cm} p{4cm} p{4cm} p{4cm}}
\caption{Multimodal fusion strategies in neuroscience FMs} \label{tab:multimodal_fusion} \\
\toprule
\textbf{Fusion Level} & \textbf{Fusion Mechanism} & \textbf{Fusion Objectives} & \textbf{Model Examples} \\
\midrule
\endfirsthead
\caption[]{(continued)} \\
\toprule
\textbf{Fusion Level} & \textbf{Fusion Mechanism} & \textbf{Fusion Objectives} & \textbf{Model Examples} \\
\midrule
\endhead
\bottomrule
\multicolumn{4}{r}{Continued on next page} \\
\endfoot
\bottomrule
\endlastfoot
\textbf{Early Fusion} & Token-level or embedding concatenation; joint encoder input & Cross-modal representation learning; Sleep staging & BENDR~\cite{kostas2021bendr}, Wei et al.~\cite{wei2025brain} \\
\textbf{Mid Fusion} & Shared attention layers; cross-token interactions across modalities & Multimodal decoding; visual reconstruction & BrainCLIP~\cite{ma2025brainclip}, MindBridge~\cite{wang2024mindbridge} \\
\textbf{Late Fusion} & Separate encoders with decision-level combination (e.g., ensemble or MLP) & Alzheimer's diagnosis; Tumor segmentation & SHADE-AD~\cite{fu2025shade}, HybridTransNet~\cite{wu2024llm} \\
\textbf{Prompt-based Fusion} & Text prompts guiding cross-modal attention or LLM generation & Clinical reasoning; psychiatric assessment & DECT~\cite{mo2025dect}, Mental-LLM~\cite{xu2024mental} \\
\textbf{Contrastive Alignment} & Cross-modal contrastive learning with shared embedding space & Semantic alignment; zero-shot generalization & NeSyGPT~\cite{cunnington2024rolefoundationmodelsneurosymbolic}, BrainCLIP~\cite{ma2025brainclip} \\
\textbf{Hierarchical Fusion} & Multi-stage modality integration (e.g., audio→video→text) & Human behavior prediction; multimodal psychiatric diagnosis & VS-LLM~\cite{wu2024vs}, Centaur~\cite{binz2024centaur} \\
\end{longtable}
}

A persistent methodological gap concerns temporal alignment across heterogeneous signals. fMRI, EEG, and MEG operate at fundamentally different temporal resolutions (seconds vs. milliseconds), creating synchronization challenges that hierarchical attention mechanisms purport to address. However, technical feasibility does not guarantee biological validity. Models that successfully align mismatched temporal scales during training may learn to exploit dataset-specific artifacts, such as stimulus timing regularities or preprocessing-induced correlations, rather than capturing genuine cross-modal neural dynamics. The distinction between fitting training data and learning generalizable neuroscience principles remains underexplored.

\subsubsection{Generative paradigms in neural reconstruction}
The structural coherence of these model categories, particularly Modality-Specific and Task-Oriented systems, is also influenced by the choice of generative architecture when reconstruction or synthesis is involved. Earlier neuroscience applications frequently adopted variational autoencoders and generative adversarial networks, which respectively offered structured latent variable modeling and high-fidelity image generation. However, in high-dimensional neural decoding settings, these paradigms exhibit structural limitations. Variational autoencoders rely on predefined latent priors and reconstruction objectives that may encourage overly smooth outputs and limit fine-grained detail recovery in perceptual reconstruction tasks~\cite{kingma2013auto, rezende2014stochastic}. Generative adversarial networks, although capable of producing sharp images, are known to be sensitive to adversarial optimization dynamics and may experience mode collapse or instability when conditioned on sparse and noisy neural signals~\cite{goodfellow2014generative, arjovsky2017wasserstein}.

More recently, diffusion models have been increasingly adopted in neuroimaging reconstruction and brain decoding tasks~\cite{ozcelik2023natural, pinaya2022brain}. Their iterative denoising formulation, which gradually transforms noise into structured outputs, allows multi-stage refinement from coarse semantic alignment to finer structural detail~\cite{ho2020denoising}. Because training is based on a noise prediction objective rather than adversarial competition, diffusion-based models tend to demonstrate improved optimization stability under weak or noisy conditioning. Autoregressive generation, by construction, factorizes the joint distribution sequentially, a design that has been linked to exposure bias and error accumulation in long-horizon sampling~\cite{bengio2015scheduled}. In contrast, diffusion models optimize a denoising objective that generates samples through iterative refinement rather than left-to-right decoding~\cite{ho2020denoising}, thereby avoiding explicit sequential dependency during generation.

While diffusion-based approaches have demonstrated practical advantages in reconstruction fidelity and optimization stability, these gains largely pertain to distributional modeling under controlled experimental conditions. They do not by themselves address longstanding challenges in neuroscience applications, including biological interpretability, robustness across acquisition sites, and the requirements of clinical deployment. In this sense, the increasing use of diffusion architectures in brain decoding studies appears to reflect practical modeling considerations in high-dimensional generative tasks, whereas their translational relevance remains to be established through broader validation.

\subsection{Training methodologies and learning paradigms}

Self-supervised learning (SSL) has emerged as the dominant pretraining strategy for neuroscience FMs, justified by the scarcity of labeled neural data~\cite{lecun2021self}. Table~\ref{tab:ssl_methods} summarizes SSL approaches adapted to neural signals. Masked signal modeling reconstructs missing segments from context, encouraging representations that capture temporal dependencies. Contrastive learning maximizes agreement between augmented views of the same signal or across modalities, promoting invariance to irrelevant transformations. Temporal forecasting predicts future states from history. Cross-modal alignment learns shared embeddings across data types.

{\footnotesize
\setlength{\tabcolsep}{4pt}
\begin{longtable}{p{3cm} p{4.2cm} p{4.2cm} p{3.5cm}}
\caption{Self-supervised learning methods for neural representation learning} \label{tab:ssl_methods} \\
\toprule
\textbf{SSL Method} & \textbf{Description} & \textbf{Key Benefits} & \textbf{Model Examples} \\
\midrule
\endfirsthead
\caption[]{(continued)} \\
\toprule
\textbf{SSL Method} & \textbf{Description} & \textbf{Key Benefits} & \textbf{Representative Models} \\
\midrule
\endhead
\bottomrule
\multicolumn{4}{r}{Continued on next page} \\
\endfoot
\bottomrule
\endlastfoot
\textbf{Masked Signal Modeling} & Predict missing segments of neural recordings using surrounding context & Context-aware embeddings, Long-range dependency modeling & BENDR~\cite{kostas2021bendr}, EEGFormer~\cite{chen2024eegformer} \\
\textbf{Contrastive Learning} & Maximize agreement between semantically similar neural signals or modalities & Modality alignment, Invariant features & BrainCLIP~\cite{ma2025brainclip}, AnatCL~\cite{barbano2024anatomical} \\
\textbf{Masked Patch Prediction} & Reconstruct missing image patches from spatial brain volumes & Spatial representation, Anatomical priors & BrainSegFounder~\cite{cox2024brainsegfounder}, FM-CT~\cite{zhu20253d} \\
\textbf{Temporal Forecasting} & Predict future neural signals from past time steps & Predictive dynamics, Temporal generalization & BRANT~\cite{zhang2023brant}, BrainLM~\cite{caro2023brainlm} \\
\textbf{Cross-modal Alignment} & Learn shared embeddings across different input modalities & Modality fusion, Semantic grounding & MindBridge~\cite{wang2024mindbridge}, BrainCLIP~\cite{ma2025brainclip} \\
\textbf{Sequential Modeling} & Capture patterns in sequential neural data & Time-series modeling, Sequential decoding & CBraMod~\cite{wang2024cbramod}, EEGFormer~\cite{chen2024eegformer} \\
\end{longtable}
}

The core assumption underlying SSL in neuroscience is that pretext tasks designed for unlabeled data will yield representations useful for downstream supervised tasks. This assumption requires empirical validation beyond correlation with benchmark performance. BENDR uses contrastive loss to distinguish temporally adjacent versus distant EEG segments~\cite{kostas2021bendr}, learning "context-aware representations" that improve sleep staging accuracy. Whether these representations capture physiologically meaningful brain states or merely statistical regularities in the training corpus is unclear. EEGFormer incorporates masked sequence modeling~\cite{chen2024eegformer}, but the biological interpretation of reconstructed masked segments remains unaddressed. BrainSegFounder combines vision transformers with patch prediction for brain segmentation~\cite{cox2024brainsegfounder}, demonstrating performance gains that could reflect either genuine anatomical pattern learning or overfitting to dataset-specific image characteristics.

A critical gap in SSL evaluation methodology is the absence of systematic failure analysis. When do SSL-pretrained models fail to generalize? Under what conditions do learned representations degrade? Most studies report aggregate performance metrics without dissecting sources of error, whether failures arise from model limitations, data quality issues, biological variability, or task misspecification. This omission limits practical utility: clinicians deploying AI systems need to understand failure modes, not just average accuracies.

\subsubsection{Model development and deployment pipeline}
\begin{figure}
    \centering
    \includegraphics[width=0.95\linewidth]{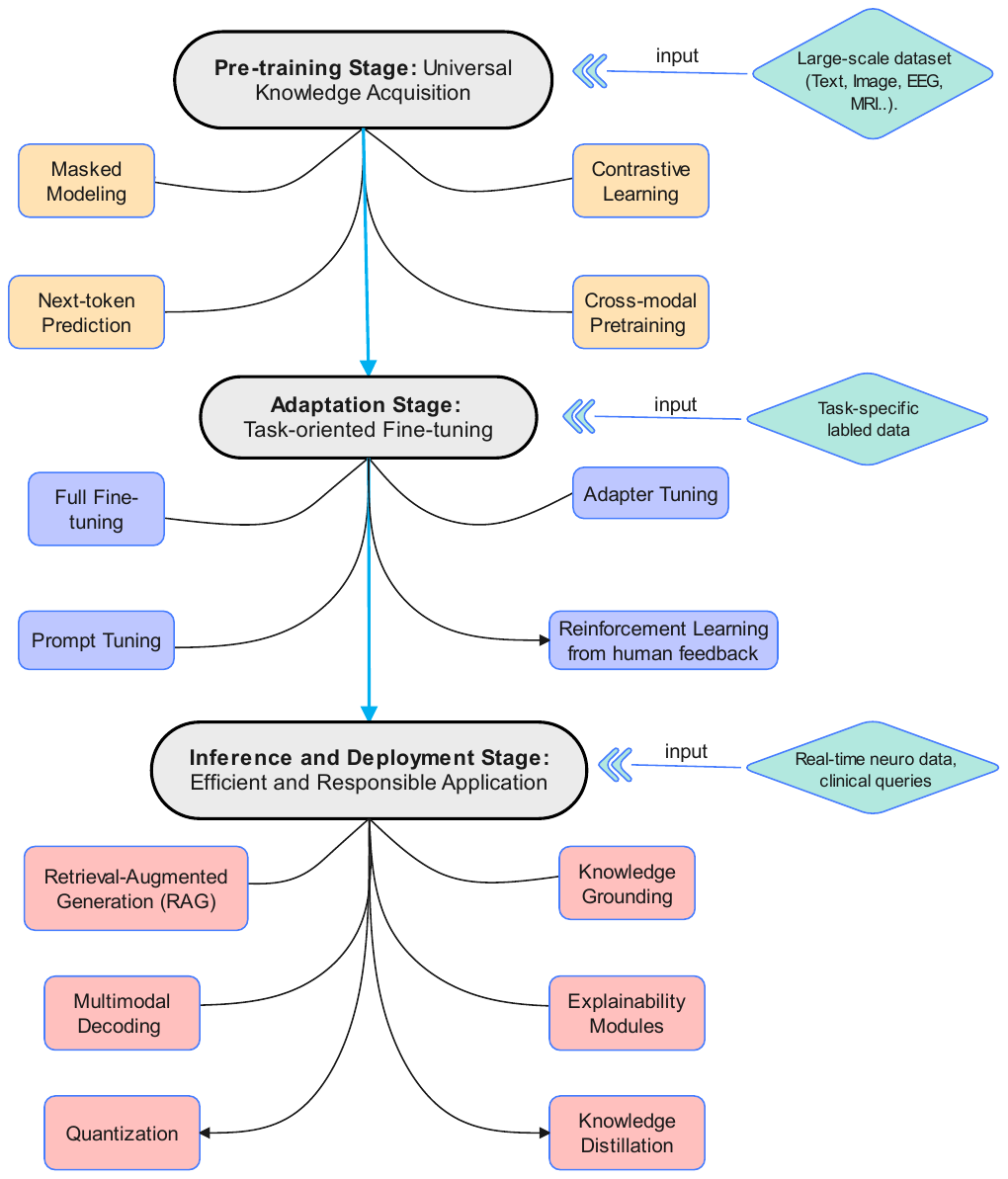}
    \captionof{figure}{Basic development pipeline of large-scale AI models}
    \label{fig:tr_ppl}
\end{figure}

The standard FM training pipeline comprises three stages: pretraining on large unlabeled datasets, adaptation to specific tasks via fine-tuning or parameter-efficient methods, and deployment for inference~\cite{bommasani2021opportunities}. Fig.~\ref{fig:tr_ppl} illustrates this workflow.

% \begin{figure}[!htbp]
%     \centering
%     \setlength{\fboxrule}{1pt}
%     \includegraphics[width=0.8\linewidth,trim= 100 0 100 10,clip]{figures/tr_ppl.pdf}
%     \caption{Basic pipeline of Large-Scale AI Models}
%     \label{fig:tr_ppl}
% \end{figure}

\textbf{Pretraining stage:} Models acquire general knowledge from diverse unlabeled data, including text corpora, image-text pairs, biological signals~\cite{bommasani2021opportunities, radford2021learning}. Data diversity purportedly enables discovery of universal patterns~\cite{qiu2020pre}, though what constitutes "diversity" in neuroscience contexts lacks operational definition. Is diversity measured by number of subjects, recording sites, experimental paradigms, or population demographics? The conflation of dataset size with diversity obscures whether models learn robust neuroscience principles or dataset-specific correlations.

\textbf{Adaptation stage:} Pretrained models are fine-tuned on labeled task-specific data, often using parameter-efficient techniques such as adapter tuning, prompt tuning, or Low-Rank Adaptation (LoRA) to reduce computational costs~\cite{houlsby2019parameter, lester2021power, hu2022lora}. This stage is critical for neuroscience applications given limited labeled data availability~\cite{brown2020language}. However, the theoretical justification for why pretraining on unrelated data (e.g., natural images) should improve neuroscience task performance remains underdeveloped. Transfer learning assumptions, that low-level features learned on source tasks transfer to target domains, may not hold when source and target involve fundamentally different data distributions and semantics. A critical yet underreported phenomenon is negative transfer, in which inductive biases learned in the source domain (e.g., general computer vision) conflict with the intrinsic spatiotemporal properties of neural signals. When the domain shift is extreme, pre-trained weights may misguide the model, leading to worse performance than training from scratch, especially in tasks that require high-frequency neural dynamic interpretation.

\textbf{Inference and deployment stage:} Real-world deployment requires not only computational efficiency but also safety, interpretability, and ethical compliance. Retrieval-augmented generation (RAG) mitigates the limitations of purely parametric knowledge by grounding model outputs in external knowledge bases~\cite{lewis2020retrieval}. However, clinical neuroscience applications face additional barriers, including regulatory approval pathways, liability for incorrect predictions, integration with established clinical workflows, and the management of edge cases that are absent from training data. Most neuroscience foundation model studies remain at the stage of methodological adaptation, with limited attention to these deployment requirements. As a result, a substantial gap persists between research prototypes and clinically validated systems.

\subsection{Methodological considerations for neuroscience foundation models}

\subsubsection{Domain-specific statistical and generalization constraints}
Neural data presents challenges distinct from natural language or computer vision domains. Signals exhibit high temporal resolution, multi-channel spatial organization, non-stationary dynamics, and substantial inter-subject variability~\cite{makeig2009linking, michel2004eeg}. These properties create methodological bottlenecks for AI model development.

\textbf{Cross-subject and cross-session generalization:} Inter-subject variability arises from anatomical differences, cognitive strategies, and individual neural characteristics; inter-session variability stems from electrode placement shifts, equipment differences, and state fluctuations~\cite{jayaram2016transfer}. Framing this as a domain adaptation problem that learns representations robust to irrelevant variations while preserving task-relevant information is conceptually appealing but operationally underspecified. What constitutes "irrelevant" versus "task-relevant" variation is context-dependent and often unknown a priori. Models that achieve high within-subject accuracy may fail catastrophically on new subjects if they exploit subject-specific idiosyncrasies rather than learning generalizable neural-behavior mappings. Systematic evaluation of cross-subject generalization requires held-out subject testing, yet many studies report cross-validation schemes that leak subject-specific information across folds.

\textbf{Temporal dynamics and multi-scale dependencies:} Neural signals span multiple timescales, from millisecond action potentials to minute-long cognitive states, requiring models that capture multi-scale temporal dependencies efficiently~\cite{tay2021scale}. Hierarchical modeling and efficient attention mechanisms address this computationally, but whether learned temporal representations align with known neurophysiological timescales remains unvalidated. Models may learn to predict behavioral outcomes from neural data without capturing the underlying temporal dynamics that neuroscientists seek to understand, creating a tension between predictive utility and mechanistic insight.

\subsubsection{Tokenization and representation design for neural signals}
Converting continuous neural signals to discrete tokens for transformer processing is nontrivial. Natural language provides semantically meaningful units, whereas neural signals require engineered tokenization schemes that balance spatiotemporal resolution with computational efficiency. For EEG and MEG, fixed-duration time windows serve as tokens, with window length trading temporal detail against computational cost. Multi-channel recordings necessitate spatial encoding via channel-wise processing combined with attention mechanisms or graph neural networks modeling electrode connectivity~\cite{li2021braingnn}. For fMRI, voxel-wise or region-of-interest tokenization compresses high-dimensional spatial data into compact representations~\cite{thomas2022self}. The critical question is whether these tokenization choices are biologically principled or merely convenient. If token boundaries do not align with neural event structures such as oscillatory cycles, event-related potentials, or functional network reconfigurations, models may fragment meaningful patterns or merge unrelated activity. Optimizing tokenization therefore requires domain knowledge about relevant spatiotemporal scales for specific neuroscience questions, which is often unavailable when designing general-purpose FMs. This creates a fundamental tension between task-agnostic tokenization for general models and task-specific representations required for neuroscience applications.

\subsubsection{Mechanistic interpretability and biological plausibility} 
Interpretability in neuroscience AI extends beyond standard explainable AI methods and requires alignment with neurophysiological mechanisms. Learned representations should correspond to known functional networks, respect anatomical connectivity constraints, and exhibit temporal dynamics consistent with neural processes~\cite{rudin2019stop}. Evaluation frameworks must therefore assess not only predictive performance but also biological plausibility. Yet biological plausibility is difficult to operationalize. It remains unclear whether models should reproduce brain-like computations at the mechanistic level or generate brain-like outputs. The former requires neurally constrained architectures, whereas the latter permits black-box function approximation. Most neuroscience FMs prioritize predictive accuracy without imposing biological constraints and only analyze learned representations post hoc for neural correspondence. This reverses the scientific priority, because if the objective is to understand neural mechanisms, biologically constrained models should be specified a priori rather than identified after training. As a result, the current paradigm risks producing high-performing yet scientifically opaque systems. This distinction is critical because models optimized solely for prediction may not provide the mechanistic explanations required for hypothesis-driven neuroscience, and therefore biological interpretability should be considered an explicit evaluation dimension in the development and assessment of neuroscience FMs.

\begin{figure}\centering
\includegraphics[width=0.95\linewidth]{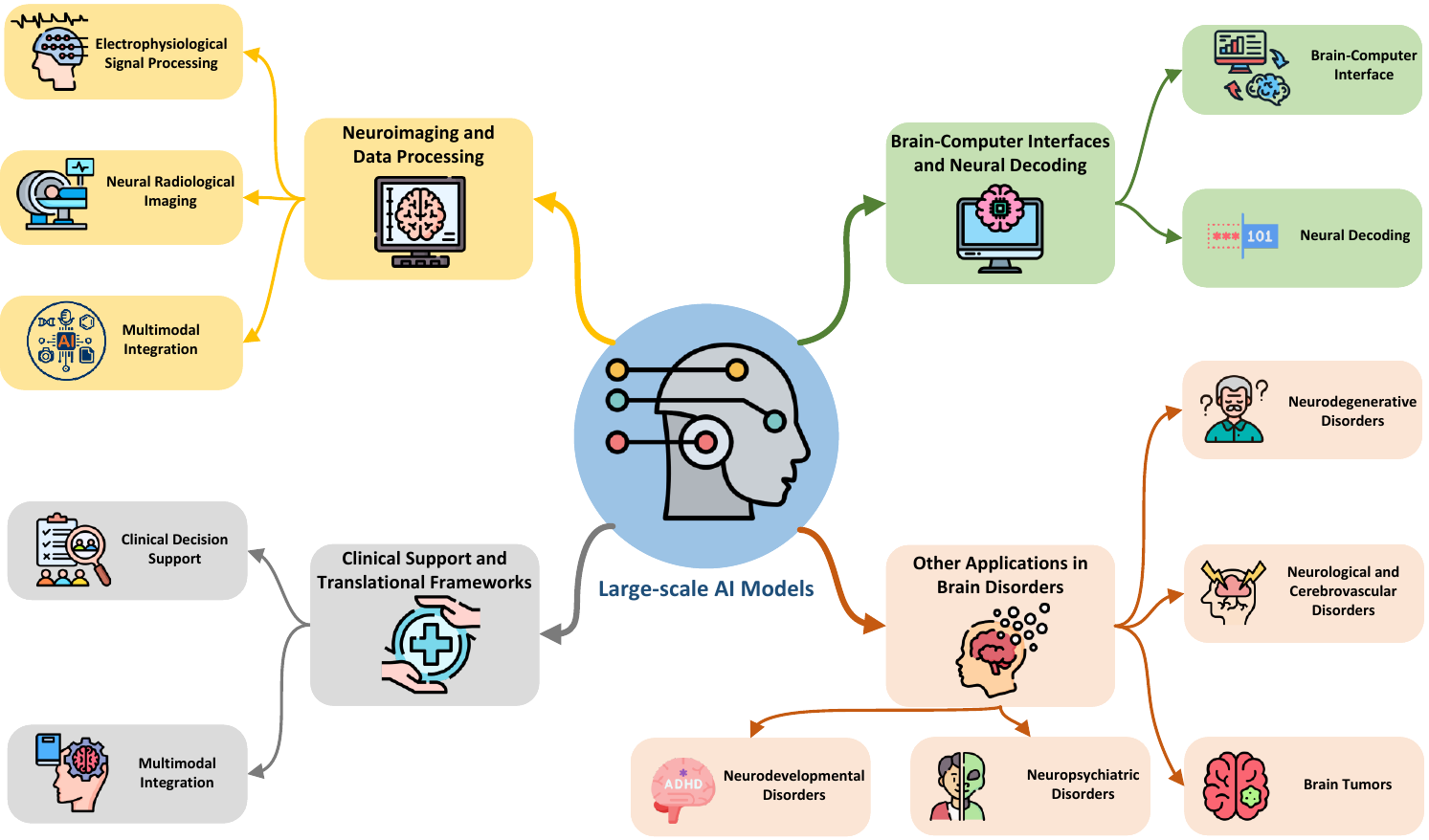}
\captionof{figure}{Overview of large-scale AI model applications in neuroscience. Four major application domains are shown, demonstrating the bidirectional relationship between neuroscience and AI development and identifying key research challenges.}
\label{fig:app_paradigm}
\end{figure}

\subsubsection{Dataset heterogeneity and validation reliability}
FM effectiveness relies on high-quality, standardized datasets that support large-scale training and cross-study evaluation. Neural data acquisition is governed by numerous technical factors, including recording hardware, experimental design, preprocessing pipelines, and artifact rejection strategies, that directly affect data quality and inter-study comparability~\cite{bigdely2015prep}. Standardization initiatives such as the Brain Imaging Data Structure (BIDS)~\cite{gorgolewski2016brain} provide important organizational frameworks, yet harmonization across recording modalities and acquisition sites remains incomplete. More fundamentally, standardization alone does not eliminate site-specific biases arising from scanner characteristics, operator expertise, and population sampling. Evidence from multi-site consortia shows that models trained on pooled datasets may encode site identity rather than clinical phenotypes and thereby achieve inflated performance by exploiting acquisition-related artifacts.

Addressing these effects requires rigorous cross-site validation, stratified evaluation, and adversarial robustness testing, methodological practices that remain uncommon in neuroscience FM research. Without systematic cross-site assessment, claims of generalization remain weak. Collectively, these limitations constrain the translational potential of large-scale AI models in neuroscience and underscore the gap between architectural capacity and clinically validated utility. The following sections examine domain-specific applications to assess whether current systems successfully address these methodological challenges.

\section{Neuroscience and neurology applications}\label{sec:applications}
Large-scale AI models are applied across several neuroscience areas, as shown in Fig.~\ref{fig:app_paradigm}, and can systematically be grouped into neuroimaging and data processing, brain-computer interfaces, genomic modeling, and clinical translation. These applications capitalize on the ability of large-scale AI models to decipher intricate neural patterns and uncover underlying laws of brain organization and function. Each group uses different architectures and methods, ranging from FM pretraining from multimodal neural data to fine-tuning with interpretability frameworks and tackling specific neuroscience research and clinical practice challenges. These methods build computational foundations of understanding neural mechanisms and permit applications in diagnostics, therapeutics, and neurotechnology across laboratory and clinical applications.

\subsection{Neuroimaging and data processing}
Large-scale AI models have expanded the methodological toolkit for neuroimaging by enabling reusable representations for anatomical volumes, functional time series, and derived connectomes. However, evidence for robustness and generalizability remains uneven across modalities because datasets, preprocessing choices, and evaluation protocols differ substantially across studies~\cite{dinsdale2022challenges}. Moreover, clinical translation faces persistent barriers in regulatory approval, workflow integration, and performance validation under real-world deployment conditions~\cite{leming2023challenges}.

\subsubsection{Neural radiological imaging models} 
FMs for radiological neuroimaging span functional MRI, structural MRI, and head CT. Performance claims are tightly coupled to the representational choices used at training time, including voxel-wise volumes, atlas-parcellated time series, and connectivity graphs. These design decisions can improve scalability and transfer in experimental evaluations, but they also impose boundaries that matter for scientific interpretation. For example, atlas-based fMRI models operate on preprocessed parcel time series and therefore do not directly model neuronal generators, while graph-based models inherit assumptions embedded in the connectivity construction procedure. Across functional and structural neuroimaging, large-scale models converge on three core design dimensions: (1) representational resolution, ranging from voxel-wise 3D volumes to atlas-based parcellated time series and connectivity graphs; (2) pretraining objectives, which primarily utilize masked autoencoding for learning intrinsic dynamics or contrastive alignment with vision-language embeddings for semantic decoding; and (3) adaptation strategies that balance subject-specific linear mapping, often implemented via ridge regression, with shared non-linear backbones to improve data efficiency.

\textbf{Functional MRI FMs:} 
Representation learning for fMRI is increasingly framed as self-supervised modeling of parcellated dynamics followed by task-specific adaptation. BrainLM~\cite{caro2023brainlm} adopts a masked autoencoding strategy to learn embeddings from 6,700 hours of fMRI recordings across UK Biobank and Human Connectome Project datasets. The model demonstrates generalization to held-out data ($R^2$=0.464 on UK Biobank, $R^2$=0.278 on HCP) and outperforms baselines in predicting clinical variables through fine-tuning. However, the performance gap between in-distribution and out-of-distribution data ($\Delta R^2$=0.186) suggests learned representations may capture dataset-specific artifacts alongside neural dynamics. While fine-tuned predictions show statistical significance, modest effect sizes ($R^2$=0.098 for future state prediction) indicate substantial unexplained variance in brain-behavior mappings. These retrospective associations within specific preprocessing pipelines (AAL-424 parcellation) do not establish mechanistic biomarkers and require prospective validation in independent clinical datasets.

A distinct research direction focuses on stimulus decoding systems in which fMRI is mapped to images through a pipeline that combines neural encoding, embedding alignment, and generative modeling. MindEye integrates contrastive retrieval with diffusion-based reconstruction, achieving approximately 93.6\% and 90.1\% accuracy for image and brain retrieval, respectively (averaged across four subjects) on the Natural Scenes Dataset~\cite{scotti2023reconstructing}, and demonstrates large-scale retrieval from LAION-5B's five billion images. However, these results are achieved under highly controlled conditions (using 7T fMRI, about 40 hours of scanning per subject, subject-specific models trained on about 25,000 samples) and do not extend to general-purpose ``brain-to-image'' capabilities in unconstrained settings. Clinical deployment requires addressing scanner heterogeneity across sites, protocol standardization, regulatory approval pathways, and prospective performance validation, none of which have been demonstrated for these methods.

MindEye2 addresses the data efficiency challenge through cross-subject pretraining, enabling high-quality reconstructions from minimal per-subject data~\cite{scotti2024mindeye2}. The model is pretrained on fMRI data from seven subjects in the Natural Scenes Dataset. After pretraining, the model can be fine-tuned on a new subject using only 1 hour of scanning data, achieving reconstruction quality comparable to previous single-subject approaches trained on 40 hours of data. This 40$\times$ reduction in required scanning time represents a significant advance toward practical applicability, reducing the barrier from dozens of MRI sessions to a single visit. The approach employs subject-specific linear mappings via ridge regression to align individual brain patterns to a shared 4096-dimensional latent space, followed by shared non-linear mappings through an MLP backbone and diffusion prior to OpenCLIP ViT-bigG/14 image space. Key remaining challenges include generalization across diverse image distributions beyond natural scenes, reliance on large-scale pretraining datasets, and robustness to variations in fMRI acquisition protocols and scanner hardware.

Several diffusion-based systems aim to improve fidelity and controllability by strengthening pretraining or structured conditioning. MinD-Vis~\cite{chen2023seeing} combines sparse-coded masked brain modeling with a double-conditioned latent diffusion formulation. The method pretrains on 136,000 fMRI segments totaling 340 hours from the HCP, GOD, and BOLD5000 datasets. On 100-way semantic classification, MinD-Vis achieves 66\% relative improvement over prior state-of-the-art, and on image generation quality measured by Frechet Inception Distance, it achieves 41\% relative improvement. The framework uses a two-stage design. First, it learns generalizable fMRI representations via masked modeling inspired by sparse coding in the visual cortex. Second, it integrates these representations with latent diffusion models through dual conditioning mechanisms. This approach enables reconstruction of semantically accurate images from brain recordings using limited paired annotations. These performance gains are measured within the study's evaluation protocol. Cross-paper comparisons remain difficult because reconstruction benchmarks, stimulus sets, and similarity metrics vary considerably across the field. Another related work has explored alternative conditioning mechanisms and strategies for improving robustness across individuals. Brain-Diffuser introduces a two-stage diffusion-based framework that combines ROI-informed analyses with multimodal CLIP-aligned semantic conditioning, using coarse structural reconstructions to guide subsequent image refinement~\cite{ozcelik2023natural}. While this design enables joint recovery of global layout and high-level semantics, its conditioning relies on predicted intermediate representations and remains sensitive to dataset-specific priors. NeuroPictor addresses inter-subject variability through multi-individual pretraining and explicit separation of semantic guidance and low-level structural modulation within the diffusion process~\cite{huo2024neuropictor}. By learning a shared fMRI latent space and refining subject-specific mappings, the approach improves reconstruction consistency under controlled experimental settings. However, the reported gains primarily reflect improved alignment within benchmark datasets rather than demonstrated generalization across recording protocols or task contexts. Across diffusion-based methods, performance gains are shown within each study's experimental setup, using evaluation metrics and stimulus sets tailored to that particular approach. Improvements in reconstruction quality or semantic alignment reflect effective representation learning under controlled conditions rather than directly comparable progress across different studies. None of these systems has been tested for robustness across different scanning protocols, task designs, or real-time constraints. Claims about brain-computer interface readiness or clinical applicability lack support from prospective validation, regulatory review, or deployment studies, and go beyond what the current evidence shows.

Beyond stimulus decoding, several studies have focused on supervised inference using brain networks derived from fMRI. The Brain Network Transformer models functional connectomes as fully connected graphs and applies self-attention to learn graph-level representations that are predictive of demographic or clinical variables~\cite{kan2022brain}. A key contribution of this work is adapting Transformer architectures to dense, ordered brain networks by using connection profiles as node features and by designing a clustering-based readout that captures modular structure without relying on predefined functional labels. While this enables effective discrimination across subjects, the learned attention patterns primarily reflect task-relevant statistical associations in functional connectivity and are not constrained to represent directed or mechanistic interactions. Hierarchical spatio-temporal modeling offers a complementary perspective by emphasizing temporal organization in fMRI-derived connectivity rather than static summaries~\cite{wei2025hierarchical}. By separating spatial and temporal processing and aggregating connectivity information across multiple scales, these models capture dynamic patterns that improve prediction performance on classification and regression tasks. Their contribution lies in demonstrating that temporal structure in functional connectivity carries additional discriminative information beyond static networks. However, the modeled dynamics describe correlations evolving over time rather than latent neural states governed by causal transitions. In these approaches, inference is performed on representations derived from functional connectivity, and model outputs are optimized for predictive accuracy under supervised objectives. Without explicit causal assumptions or validation against controlled perturbations, such models do not establish effective connectivity or recover directed neural interactions, limiting their interpretability in terms of underlying neural mechanisms.

In parallel with these task-specific advances, broader pretraining strategies have been proposed to improve transfer across datasets, though their implications for clinical generalization remain less clear. BrainGFM treats each scan as a graph representation and learns embeddings with graph-aware encodings and Transformer components~\cite{wei2025brain}. Pretraining spans 27 datasets covering 25 neurological disorders with over 25{,}000 subjects and more than 60{,}000 fMRI scans across eight atlases. This breadth enables improved transfer across parcellation schemes in the tested evaluations. The use of few-shot adaptation and language-prompted learning as an adaptation interface does not imply that the model generalizes to clinical heterogeneity without careful task definition and dataset-specific validation. Prospective evaluation in new clinical cohorts with pre-specified endpoints remains absent.

\textbf{Structural MRI FMs:} 
While functional MRI FMs target dynamic brain states and stimulus-response relationships, structural MRI FMs emphasize anatomical representation learning for segmentation and morphometric analysis. The shift from temporal dynamics to spatial structure introduces distinct technical challenges in handling anatomical variability and pathological heterogeneity.

Structural MRI FMs primarily focus on learning anatomical representations to support segmentation and related downstream tasks. BrainSegFounder adopts a two-stage self-supervised pretraining strategy, first encoding normative brain anatomy from 41{,}400 participants in the UK Biobank and subsequently adapting these representations using disease-specific datasets. Fine-tuning is evaluated on lesion and tumor segmentation benchmarks, including ATLAS v2.0 with 655 MRIs collected from 44 datasets and BraTS 2021 with 1{,}251 multi-modal MRI volumes~\cite{cox2024brainsegfounder}. Quantitative results demonstrate consistent improvements in volumetric segmentation accuracy across datasets included in the study. The large-scale anatomical pretraining stage promotes the learning of global structural context and spatial regularities in 3D MRI, which provides a stable inductive bias for downstream segmentation. Subsequent adaptation refines these representations toward pathology-specific geometry and spatial distribution, enabling effective transfer with limited labeled data. Compared with prompt-driven adaptation approaches such as MedSAM~\cite{ma2024segment}, which emphasize task-time interaction, BrainSegFounder prioritizes offline representation learning to support task-agnostic 3D feature reuse. However, the learned representations are shaped predominantly by healthy population statistics and a narrow set of target pathologies. Generalization to unseen acquisition protocols, scanner variability, or rare disease phenotypes remains insufficiently characterized.

BrainIAC frames generalization as a pretraining and adaptation strategy across heterogeneous downstream tasks~\cite{tak2024foundation}. The model is pretrained using self-supervised contrastive learning on 32{,}000 brain MRI scans aggregated from 35 datasets and is evaluated on a total of 48{,}519 scans spanning multiple classification and regression problems. Across curated benchmarks, a single shared backbone supports reuse across tasks with varying objectives and data availability. Rather than emphasizing anatomy-specific priors, the contrastive objective encourages the encoding of modality-agnostic structural and intensity patterns that recur across brain MRI acquisitions. This design facilitates flexible task-specific adaptation but is assessed primarily under retrospective and controlled evaluation settings. As a result, claims regarding clinical readiness remain premature in the absence of prospective studies examining deployment robustness, workflow integration, and failure modes in real-world clinical populations.

When manual annotations are limited, weakly supervised anatomical representation learning offers an alternative strategy for model pretraining. AnatCL introduces a contrastive framework that incorporates anatomical descriptors and age-aware signals as proxy supervision, and evaluates the learned representations across multiple diagnostic classification tasks and continuous assessment score prediction tasks~\cite{barbano2024anatomical}. Performance gains in these settings suggest that the learned features capture subtle and spatially distributed anatomical variation that may be difficult to encode through fully supervised objectives alone. At the same time, the effectiveness of this approach is closely tied to the fidelity of the proxy supervision. Anatomical measures and age-related signals may reflect population-level trends rather than disease-specific effects, complicating the separation of pathological variation from confounding factors such as scanner site, demographic composition, and acquisition protocol differences.

\textbf{CT FMs:} 
Head CT FMs target time-critical triage scenarios, though current evidence comes primarily from retrospective benchmarks. Two recent approaches illustrate contrasting design strategies. DeepCNTD-Net constructs task-aligned volumetric representations by combining LLM-assisted multi-label annotation with domain-specific pretraining for hemorrhage subtype segmentation and brain anatomy parcellation, which are subsequently fused through multimodal fine-tuning for comprehensive neuro-trauma detection~\cite{yoo2025non}. This design enables simultaneous detection of 16 clinically relevant findings and yields consistent improvements over generic CT FMs in retrospective evaluations, highlighting the value of explicitly encoding neuro-anatomical and pathological structure for triage-oriented tasks. FM-HCT, by contrast, adopts a representation-centric paradigm based on large-scale self-supervised pretraining on 361{,}663 non-contrast 3D head CT scans using a customized 3D Vision Transformer, followed by fine-tuning for multi-label disease detection across internal and external benchmarks~\cite{zhu20253d}. Performance gains observed under both internal validation and cross-institutional testing indicate that data scale and architectural capacity can support transferable volumetric representations without task-specific supervision. These two approaches reflect different trade-offs. DeepCNTD-Net emphasizes structured integration of anatomy- and pathology-aware features, favoring robustness within predefined neuro-trauma triage protocols. FM-HCT relies instead on generic volumetric feature learning to support adaptation across datasets and limited-label regimes. While both models demonstrate improved detection performance relative to training from scratch in retrospective settings, their evaluations remain confined to curated benchmarks. As a result, differences in calibration behavior, failure modes, and robustness under routine clinical variability cannot be assessed from the reported experiments alone.

While the above models differ in architecture, modality, and task, they converge on a consistent set of trade-offs. Large-scale pretraining (e.g., MinD-Vis: 136,000 segments, FM-HCT: 361,663 scans, etc.) consistently improves transfer performance within curated benchmarks. However, advances in data efficiency (e.g., MindEye2: one-hour adaptation) remain confined to controlled experimental settings. A persistent gap separates retrospective benchmark performance from prospective clinical validation: no reviewed model has demonstrated effectiveness through pre-registered trials in real-world deployment settings. This gap reflects not merely insufficient validation effort, but fundamental technical challenges, scanner heterogeneity introduces systematic biases that curated benchmarks systematically exclude, protocol drift across sites creates distribution shifts that current transfer learning approaches do not handle robustly, and real-world clinical populations present comorbidities and edge cases underrepresented in research datasets.

\subsubsection{Electrophysiological signal processing models}

Foundation models for electrophysiological recordings are increasingly framed as large-scale representation learners rather than end-to-end clinical systems. Empirically, the strongest evidence to date comes from retrospective benchmarking on heterogeneous datasets, where gains often depend on preprocessing choices, label definitions, and the fine-tuning protocol~\cite{dinsdale2022challenges}. Across studies, persistent constraints include sensitivity to montage variation, nonstationarity, and artifact contamination, none of which is fully resolved by scaling alone. Moreover, clinical translation faces barriers in workflow integration, prospective validation, and regulatory approval~\cite{leming2023challenges}. Electrophysiological FMs share a consistent methodological structure organized around spatiotemporal tokenization, where continuous neural recordings are discretized into fixed-duration patches. Methodological distinctions in this domain arise from (1) spatial dependency handling, moving from fixed electrode montages to topology-agnostic attention mechanisms that decouple computation from specific channel layouts ; and (2) learning paradigms, which alternate between autoregressive next-token prediction and context-aware masked reconstruction of contiguous signal segments, or augmented by temporal-frequency decoupling.

\textbf{EEG FMs:} Early EEG FMs explored different design priorities when applying self-supervised learning to large-scale clinical recordings. Some work treats interpretability as an explicit design objective rather than an emergent property of attention mechanisms. EEGFormer adopts a vector-quantized Transformer framework and is pretrained on approximately 1.7~TB of unlabeled EEG data from the TUH corpus~\cite{chen2024eegformer}. By learning a discrete codebook over EEG patches, the model enables post hoc analysis of temporal and spatial patterns associated with downstream predictions, providing a structured basis for examining recurrent signal motifs. The discrete representations learned by the vector quantizer support inspection of token usage and n-gram statistics across channels and time, which can facilitate hypothesis generation about discriminative structure in EEG signals. However, such analyses remain descriptive in nature. While they offer insights into patterns captured by the model, they do not establish physiological causality, and their interpretability depends on modeling choices such as codebook design and the specific attribution or analysis procedures employed.

Generative or language-model-inspired training has also been explored, but the demonstrated capabilities remain task- and dataset-dependent. Neuro-GPT combines self-supervised pretraining with downstream adaptation for EEG tasks~\cite{cui2024neuro}. The model couples an EEG encoder with a decoder only GPT backbone and applies a causal masking objective to reconstruct masked EEG chunks from preceding context, rather than directly forecasting raw EEG segments. While autoregressive formulations can naturally accommodate variable-length sequences, Neuro-GPT uses fixed-length chunk sequences with zero-padding in its experiments, and the empirical evidence in this line of work is primarily based on retrospective benchmarks, and it does not directly establish robust real-time performance under clinical noise, electrode dropout, or protocol shift. The reliance on temporal dependencies may also limit applicability to tasks where critical discriminative information appears in isolated temporal windows rather than extended sequences.

Scaling and breadth of supervision are pursued more explicitly in multitask pretraining. EEGPT pretrains a 10-million-parameter transformer on a mixed multi-task EEG dataset comprising motor imagery, motor execution, steady-state visual evoked potential (SSVEP), emotion recognition, and identification paradigms, utilizing up to 58 electrodes standardized to a 10-20 system configuration~\cite{NEURIPS2024_4540d267}. The architecture employs a hierarchical structure that decouples spatial and temporal feature extraction: an encoder first integrates spatial information across electrode channels for each time segment, then a predictor captures temporal dependencies across segments. This design reduces computational complexity while maintaining flexibility for brain-computer interface (BCI) applications. Pretrained models are evaluated via linear probing across diverse downstream tasks including motor imagery classification (BCIC-2A, BCIC-2B), sleep stage detection (Sleep-EDFx), event-related potential detection (KaggleERN, PhysioP300), abnormal EEG detection (TUAB), and event-type classification (TUEV). The largest variant (101 million parameters) achieves state-of-the-art performance across multiple benchmarks, demonstrating that larger models exhibit improved accuracy following scaling laws of the form $\mathrm{ACC} = (33.6 \times N)^{0.029}$, where $N$ is the parameter count. However, cross-study comparisons remain challenging due to non-standardized evaluation protocols, and the linear-probing approach, while avoiding overfitting on small downstream datasets, may limit adaptation to tasks with substantially different channel configurations or paradigms not represented in the pretraining mixture.

Temporal modeling innovations have been proposed to better match EEG dynamics, especially for detection settings where short-lived patterns matter. FoME introduces ATLAS (Adaptive Temporal-Lateral Attention Scaling) and uses large-scale pretraining on 1.7 TB of data from 15{,}000 subjects with a 745M-parameter model trained for 1{,}096{,}000 steps~\cite{shi2024fome}. Evaluations across four downstream tasks cover classification, detection and forecasting, demonstrating feasibility for time-sensitive decision support in controlled settings. However, clinical deployment requires prospective workflow studies, real-time performance validation under clinical artifacts, and regulatory approval, none of which have been pursued for this model.

Several recent models target architectural mismatches between NLP-style transformers and multiscale neural activity. CSBrain proposes cross-scale spatiotemporal tokenization with structured sparse attention and evaluates across 11 EEG tasks and 16 datasets~\cite{zhou2025csbrain}. The tokenization strategy segments EEG into patches at multiple temporal resolutions and anatomically informed spatial groupings derived from standard electrode layouts, creating a hierarchical representation that preserves both local detail and global context. Structured sparse attention reduces computational complexity from quadratic to approximately linear in sequence length by limiting attention to within-scale and cross-scale connections. The evidence supports broad benchmarking coverage across motor imagery, visual processing, and cognitive tasks, yet the degree to which gains arise from tokenization, attention structure, or dataset-specific training details is not always separable from published summaries alone. Moreover, the hierarchical design assumes stationary electrode layouts, which may limit generalization to recordings with missing or repositioned channels.

Efficiency-oriented designs have also been explored for real-time feasibility. CodeBrain introduces a decoupled temporal-frequency tokenizer and demonstrates generalization experiments across 10 public EEG datasets~\cite{ma2025codebrain}. The decoupling strategy first applies time-domain convolutions to extract temporal features, then performs frequency-domain analysis on these representations through learned spectral basis functions, effectively combining time-locked event detection with frequency-band power estimation. This dual representation enables the model to capture both phase-locked responses (e.g., evoked potentials) and non-phase-locked phenomena (e.g., induced oscillations) within a unified framework. These results suggest that separating temporal and spectral discretization can be useful for transfer, although the computational tradeoffs and stability under montage shift still require clearer cross-site validation, and the learned spectral bases may not align with canonical frequency bands used in clinical interpretation.

Multimodal pretraining is emerging as a strategy for integrating complementary noninvasive neural recording modalities. BrainOmni performs large-scale self-supervised pretraining on 1{,}997 hours of EEG data and 656 hours of MEG data, and introduces a sensor-aware tokenizer designed to mitigate dependence on device-specific channel naming conventions and fixed sensor topologies~\cite{xiao2025brainomni}. By explicitly encoding sensor location, orientation, and type, the model learns unified representations that bridge heterogeneity across EEG and MEG acquisition systems. This design shifts the alignment problem from channel identity to shared spatiotemporal structure, enabling joint modeling of signals with distinct physical measurement characteristics. Experimental results support the feasibility of learning unified embeddings across modalities under the evaluated benchmarks. However, current evidence remains limited to the datasets and device splits considered in the study, and does not yet establish robust generalization to substantially different acquisition systems or to diverse clinical populations.

Broader attempts to reduce dependence on fixed electrode layouts include topology-agnostic and instance-adaptive pretraining strategies. LUNA emphasizes generalization across heterogeneous electrode configurations and computational efficiency by projecting variable-channel EEG into a fixed-size latent space via learned queries and cross-attention, thereby decoupling downstream computation from montage size. Pretrained on over 21{,}000 hours of EEG from the Temple University EEG Corpus (TUEG) and the Siena dataset, LUNA achieves an AUROC of 0.921 on TUAR while reducing FLOPs by up to 300$\times$ and GPU memory usage by up to 10$\times$, with consistent gains across evaluated electrode configurations~\cite{donerluna}. In contrast, CRIA addresses data heterogeneity through instance-adaptive variable-length and variable-channel coding combined with explicit cross-view interaction across temporal, spectral, and spatial representations. Under its stated experimental settings, CRIA reports a balanced accuracy of 57.02\% for multi-class EEG event classification and an AUROC of 80.03\% for anomaly detection on Temple University EEG and CHB-MIT datasets~\cite{liu2025cria}. These results indicate tangible progress toward robustness across heterogeneous subjects and electrode layouts, while also underscoring that performance remains modest for fine-grained EEG event taxonomies and complex clinical annotation schemes.

\textbf{Advanced neural recording models:}
Intracranial FMs have been developed to learn patient-agnostic structure from long recordings, but the evidence base is still largely retrospective. BRANT uses a 500M-parameter model trained with masked patch prediction on an intracranial dataset comprising 2{,}528 hours at 1{,}000 Hz (approximately 1.01 TB) from 152 participants~\cite{zhang2023brant}. The masking strategy operates on contiguous temporal patches of 2 seconds, requiring the model to reconstruct masked time–frequency representations from surrounding context. This supports the feasibility of large-scale self-supervision for sEEG representation learning, particularly for capturing recurring epileptiform patterns that appear consistently across different brain regions and subjects. However, it does not eliminate subject-specific confounds introduced by implantation strategy, pathology distribution, or recording hardware, and the learned representations may overfit to epileptiform activity patterns that dominate the clinical dataset.

Beyond learning general representations, intracranial modeling also faces challenges from sparse spatial sampling and subject-specific electrode placement. BrainStratify proposes a coarse-to-fine architecture with decoupled product quantization and evaluates performance on two sEEG datasets and one epidural ECoG dataset for speech perception and production decoding~\cite{zheng2025brainstratify}. The coarse-to-fine strategy first identifies broad functional groups across electrode contacts using low-dimensional embeddings, then refines these representations through contact-specific feature extraction that captures task-relevant local neural dynamics. Decoupled product quantization compresses representations by independently quantizing multiple latent subspaces corresponding to distinct neural dynamics, reducing memory requirements while preserving discriminative information for downstream tasks. The results demonstrate consistent improvements over established baselines in phoneme classification and word reconstruction accuracy within speech decoding tasks. However, broad claims about universal intracranial decoding should be avoided given the narrow task focus and limited dataset diversity, and the approach does not address the challenge of transferring learned representations across subjects with different implantation sites or underlying pathologies.

Cross-modality training has been explored more explicitly in BrainWave, which unifies EEG and sEEG within a single self-supervised pretraining framework~\cite{yuan2024brainwave}. The pretraining corpus comprises 13.79~TB of electrical brain recordings spanning 40{,}907 hours from 15{,}997 individuals, including 27{,}063 hours of TUH EEG from 14{,}987 subjects alongside a large-scale sEEG corpus. The model is evaluated across a broad set of downstream tasks under cross-domain and few-shot settings, covering variation across subjects, institutions, and disease subtypes. BrainWave addresses cross-modality heterogeneity by aligning signals with different sampling rates and channel configurations into a shared latent space through scale alignment and channel-agnostic modeling. Experimental results indicate that representations learned from joint pretraining can transfer across invasive and noninvasive recordings in controlled retrospective evaluations. However, these findings are from curated benchmarks. Generalization to prospective clinical settings, cross-institutional label inconsistencies, and real-world artifacts has not been evaluated.

\textbf{Specialized EEG applications:}
Scaling studies have started to examine parameter and data scaling effects in EEG FMs.  ALFEE scales to 540M parameters with 25{,}000 hours of EEG pretraining and evaluates across six downstream tasks including emotion recognition, sleep staging, and abnormality detection~\cite{xiong2025alfee}. The architecture employs depth-wise separable convolutions in early layers to process individual channels independently before cross-channel integration, balancing spatial and temporal modeling efficiency. The improvements support the practicality of large-scale EEG representation learning, while leaving open questions about calibration, failure modes, and stability under cross-site protocol shift. Notably, performance on event type classification remained substantially lower than that achieved on other detection tasks, highlighting the challenges of fine-grained EEG event labeling and the potential need for higher-quality annotations or task-specific supervision.

Masking strategies that are tuned to electrophysiological structure can matter as much as backbone choice. LaBraM adapts masked modeling to EEG temporal dynamics and demonstrates that EEG-specific design choices outperform direct transplantation of vision-style masking heuristics in its studied settings~\cite{jiang2024large}. Specifically, the model masks contiguous temporal patches during pre-training rather than individual time points, reflecting the structured temporal nature of EEG signals instead of adopting fixed or purely random masking schemes from vision models. This domain-informed masking produced measurably better transfer performance across multiple downstream EEG classification tasks compared to uniform random masking. The results support a broader point that modality-appropriate pretext design is often necessary for reliable transfer, though the optimal masking distribution may vary across recording contexts and clinical applications, and current implementations do not adaptively adjust mask properties based on signal characteristics.

Disease-focused applications should be framed as decision-support models validated on specific datasets rather than as deployable diagnostic tools. LEAD targets EEG-based Alzheimer's disease detection, pretraining on 2{,}354 subjects across multiple datasets and evaluating on an 813-subject EEG-AD dataset, with reported relative F1 score improvements of up to 9.86\% at the sample level and 9.31\% at the subject level under the stated evaluation protocols~\cite{wang2025lead}. The model incorporates graph neural network layers that model functional connectivity patterns derived from coherence and phase-locking measures between electrode pairs, hypothesizing that Alzheimer's-related degradation manifests in disrupted long-range connectivity. These results support feasibility for dataset-level classification in research settings where EEG recordings are collected under controlled protocols. However, brain-cognitive gaps, which is the discrepancy between neural signal patterns and cognitive outcomes introduce uncertainty in predictive validity, as EEG features optimized for classification accuracy may not capture the cognitive mechanisms underlying dementia progression~\cite{esmaeili2024brain}. Clinical deployment would require prospective performance validation in screening scenarios, workflow integration that addresses variable recording quality and patient compliance, and regulatory readiness for diagnostic claims that would require FDA clearance or equivalent approval.

Despite architectural diversity, EEG FMs converge on common challenges that limit clinical translation. Large-scale pretraining (e.g., FoME: 1.7 TB from 15{,}000 subjects, etc.) consistently improves transfer performance within retrospective benchmarks. However, gains are tightly coupled to preprocessing choices, electrode montages, and fine-tuning protocols that vary across studies. Persistent technical barriers include sensitivity to montage variation, nonstationarity, artifact contamination, and other challenges that scaling alone does not resolve. Moreover, prospective clinical validation remains absent. Current research lacks prospective validation in real-world clinical settings through pre-registered trials. Achieving clinical utility requires addressing key implementation barriers, including compatibility with clinical workflows, regulatory compliance, inter-institutional protocol standardization, and robustness to real-time data artifacts. This methodological synthesis reveals that while architectural diversity is high, the field is moving toward modality-agnostic encoders and hierarchical attention structures that aim to reconcile the high temporal resolution of electrophysiology with the multi-scale organization of brain networks.

\begin{figure}\centering
\includegraphics[width=0.8\linewidth]{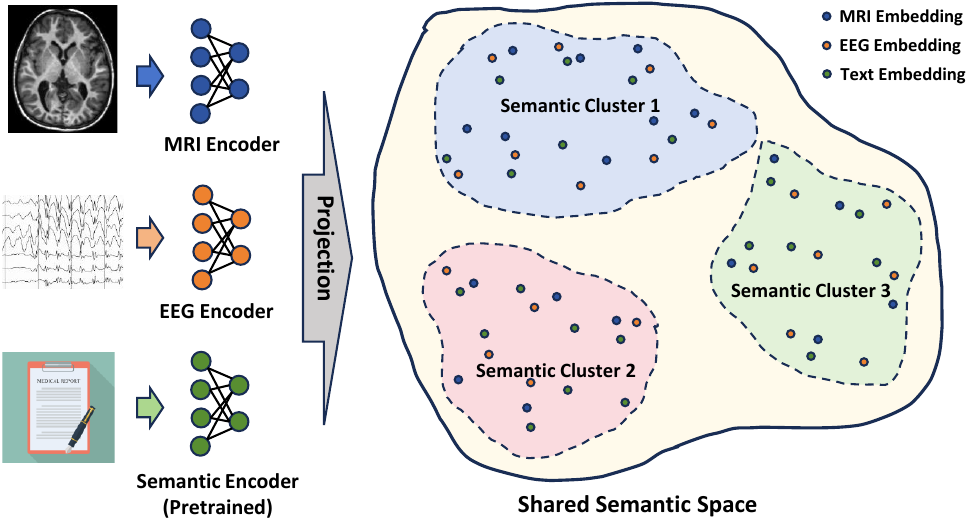}
\captionof{figure}{Representation-level multimodal alignment.Modality-specific encoders project heterogeneous data into a shared pretrained semantic space, enabling subject-invariant, cross-task decoding and shifting multimodal integration from feature fusion to alignment.}
\label{fig:mm_integration}
\end{figure}

\subsubsection{Multimodal integration approaches}

Multimodal integration in neuroscience increasingly takes the form of representation alignment across heterogeneous measurements, rather than naïvely concatenating features. This paradigm is illustrated in Fig.~\ref{fig:mm_integration}, where heterogeneous modalities are projected into a shared semantic space that supports multiple downstream decoding objectives. A recurring design choice is to map neuroimaging signals into a shared embedding space learned from large-scale naturalistic supervision, then evaluate decoding performance with identification, retrieval, or reconstruction metrics under controlled protocols. These studies provide useful computational probes for linking brain activity to perceptual or semantic structure, but most evidence remains offline and dataset-bound, with limited support for deployment claims.

\textbf{Cross-modal brain decoding:} A prominent line of work anchors brain decoding to pretrained vision--language embeddings, treating the embedding space as an intermediate representation that supports both retrieval-style evaluation and image reconstruction. MindBridge explicitly targets cross-subject decoding by aligning subject-specific fMRI signals to CLIP embeddings and employing a diffusion-based dual-channel decoder for image reconstruction~\cite{wang2024mindbridge}. Experiments are conducted on the Natural Scenes Dataset using four subjects, each with 8{,}859 training images and a shared test set of 982 images. Under this cross-subject setting, MindBridge reports an average PixCorr of 0.151 and an SSIM of 0.263 across the four subjects, together with a CLIP similarity score of 94.7\%. In this framework, the use of CLIP embeddings provides a subject-invariant semantic target, reducing dependence on idiosyncratic voxel-level patterns. The framework further examines adaptation to an unseen subject through lightweight tuning. With 500, 1{,}500, and 4{,}000 subject-specific training images, it reports PixCorr values of 0.112, 0.140, and 0.156, alongside corresponding SSIM values of 0.229, 0.250, and 0.258, respectively. These results indicate that embedding-aligned decoding can extend beyond single-subject training, while also showing that image-level reconstruction quality remains sensitive to the amount of subject-specific data available.

BrainCLIP adopts a related alignment strategy but frames evaluation more explicitly as embedding-space retrieval from fMRI, again using the Natural Scenes Dataset as a primary benchmark~\cite{ma2025brainclip}. Each participant is trained with 8{,}859 image stimuli, and retrieval is evaluated against a fixed candidate pool of 982 images at test time. Under this setting, the model demonstrates strong fMRI-to-image and fMRI-to-text retrieval performance. The study further evaluates zero-shot visual category decoding on the Generic Object Decoding dataset, where test categories are disjoint from those used during training. BrainCLIP treats CLIP embeddings as a shared semantic target that facilitates alignment across subjects and tasks, enabling retrieval and classification without explicit category-specific supervision. Taken together, these results indicate that CLIP-style semantic embeddings can serve as a workable intermediate representation for fMRI-to-vision decoding under standard candidate-set protocols. At the same time, the empirical evidence is largely limited to retrospective benchmarks with fixed stimulus sets, and the embedding target prioritizes semantic consistency over pixel-faithful reconstruction.

A complementary perspective is offered by FMs that are not trained on neural data but are evaluated as candidate representational accounts of cortical activity. BriVL is a multimodal FM pretrained on 15 million image--text pairs and assessed through neural encoding analyses rather than direct brain decoding~\cite{lu2022multimodal}. Alignment to fMRI responses is evaluated on a visual dataset comprising five subjects viewing approximately 10.2 hours of natural movie stimuli, and on a separate language dataset with eight subjects reading 384 unique sentences. Encoding models are fitted using banded ridge regression and evaluated using split-$R^2$, with voxel selection based on explained variance within the analysis pipeline. These results suggest that certain layers of multimodally trained visual and linguistic encoders capture stimulus features that correlate with cortical responses across both vision and language domains. At the same time, the study explicitly frames these findings as evidence of representational alignment rather than a mechanistic neural model. The reported encoding performance is therefore constrained by the specific stimuli, acquisition settings, and linear readout assumptions employed in the evaluation.

\textbf{Integration of language and brain function:} Recent work has begun to operationalize language-related decoding as conditional text generation from fMRI-derived features, often within a multimodal system that includes auxiliary context signals. BrainDEC utilizes an end-to-end multimodal LLM to decode spoken text from non-invasive fMRI recordings during natural French conversations~\cite{HMAMOUCHE2026103589}. This framework addresses the temporal mismatch between slow-varying BOLD signals and rapid linguistic tokens by using an Inception-based deconvolution layer that extracts multiscale temporal features and aligns them with a frozen Llama-2-7B backbone via a bipartite transformer architecture. While evaluations of human-human and human-robot interactions demonstrate the feasibility of reconstructing semantic meaning under controlled experimental designs, the generated output shows modest performance on text-overlap and semantic metrics. This suggests that while the generative priors of FMs can effectively supplement the low signal-to-noise ratio of non-invasive recordings, the practical utility of such systems remains constrained by hemodynamic latency and the challenge of maintaining decoding robustness across diverse speakers and scanning protocols.

\textbf{Structural-functional integration:} Multimodal fusion is also used for disease-related classification tasks that combine anatomical and functional summaries, with architectural choices designed to preserve modality-specific structure. MultiViT fuses 3D gray matter volume maps from structural MRI with 2D functional network connectivity matrices derived from ICA-processed resting-state fMRI, using separate ViT encoders and cross-attention for feature fusion~\cite{bi2024multimodal}. The paper evaluates on multi-site schizophrenia datasets comprising over 2,000 participants scanned with 3T scanners, with data collected across multiple sites. In the reported training and evaluation protocol, a 10-fold cross-validation is used to summarize performance. MultiViT achieves an AUC of 0.833, exceeding unimodal and multimodal baselines. The work also derives saliency-style interpretations from attention rollout and contrasts groups using a two-sample t-test, with a reported $p$-value threshold of 0.02 to highlight group-differentiating regions. These results support the claim that cross-attention fusion can improve retrospective classification performance under the study's evaluation pipeline. However, they do not establish clinical readiness, since the evidence is limited to retrospective model validation and does not report prospective deployment, regulatory evaluation, or real-world workflow testing.

{\scriptsize
\setlength{\tabcolsep}{2.8pt}
\begin{longtable}{p{2.2cm} p{3.3cm} p{2.0cm} p{2.2cm} p{3.3cm} p{2.0cm}}
    \caption{Comprehensive overview of large-scale AI models for computational neuroscience research. (PT: Pre-training; FT: Fine-tuning). The column "Multi-Center Data" categorizes the scope of training data, highlighting the prevalence of multi-site datasets in the current landscape.} \label{tab:neuroscience_models} \\
    \toprule
    \textbf{Model Name} & \textbf{Primary Application} & \textbf{Category} & \textbf{Training Method} & \textbf{Dataset Used} & \textbf{Multi-Center Data} \\
    \midrule
    \endfirsthead

    \caption[]{-- continued from previous page} \\
    \toprule
    \textbf{Model Name} & \textbf{Primary Application} & \textbf{Category} & \textbf{Training Method} & \textbf{Dataset Used} & \textbf{Multi-Center Data} \\
    \midrule
    \endhead

    \bottomrule
    \multicolumn{6}{r}{Continued on next page} \\
    \footnotemark
    \endfoot

    \bottomrule
    \endlastfoot

ADAgent~\cite{hou2025adagentllmagentalzheimers} & Alzheimer's disease analysis & System w/ LLM & LLM integration & ADNI dataset (T1-MRI, PET-FDG) & Yes \\
ALFEE~\cite{xiong2025alfee} & General EEG representation learning for multiple tasks & FM & PT + FT & Multi-task EEG datasets including clinical, emotion, sleep, and workload & Yes \\
AnatCL~\cite{barbano2024anatomical} & Anatomical FM for structural brain MRI analysis & FM & PT + FT & Multiple MRI datasets with Desikan-Killiany atlas & Yes \\
ASD-Chat~\cite{deng2024asd} & Autism spectrum disorder support & System w/ LLM & Prompt-based & Clinical intervention data & No \\
AtlasGPT~\cite{hopkins2024atlasgpt} & Neurosurgical clinical augmentation & System w/ LLM & RAG-enhanced LLM & JNSPG articles & No \\
BENDR~\cite{kostas2021bendr} & Transformer-based EEG analysis for BCI applications & FM & PT + FT & TUEG and other large EEG datasets & Yes \\
BRANT~\cite{zhang2023brant} & FM for intracranial neural signal analysis & FM & PT + FT & Large corpus of intracranial SEEG data & No \\
BrainBench~\cite{luo2024large} & Neuroscience benchmarking framework & LLM & PT + FT & Neuroscience literature & Yes \\
BrainCLIP~\cite{ma2025brainclip} & Task-agnostic fMRI brain decoding with CLIP integration & FM & PT + FT & fMRI data with visual stimuli & Yes \\
BrainDEC~\cite{HMAMOUCHE2026103589} & Non-invasive decoding of spoken text from fMRI & Framework w/ LLM & PT + FT & Custom corpus with fMRI recordings & Yes \\
Brain-Diffuser~\cite{ozcelik2023natural} & fMRI-to-image reconstruction using generative diffusion & FM & PT + FT & Natural Scenes Dataset (NSD) & No \\
BrainGFM~\cite{wei2025brain} & Unified framework for large-scale fMRI pre-training & FM & PT + FT & 27 datasets, 25,000+ subjects, 60,000 fMRI scans & Yes \\
BrainGPT~\cite{li2025towards} & Neurosurgical applications & LLM & PT + FT & 1.3B tokens (neuroscience literature) & No \\
BrainIAC~\cite{tak2024foundation} & General-purpose FM for structural brain MRI & FM & PT + FT & 48,519 brain MRIs from 35 datasets & Yes \\
BrainLM~\cite{caro2023brainlm} & FM for brain activity dynamics prediction & FM & PT + FT & UK Biobank, HCP (6,700 hours fMRI recordings) & Yes \\
BrainOmni~\cite{xiao2025brainomni} & First FM for unified EEG and MEG analysis & FM & PT + FT & 1,997 hours EEG + 656 hours MEG & Yes \\
BrainSegFounder~\cite{cox2024brainsegfounder} & 3D FM for multimodal neuroimage segmentation & FM & PT + FT & UK Biobank (41,400 participants), BraTS, ATLAS v2.0 & Yes \\
BrainStratify~\cite{zheng2025brainstratify} & FM for invasive and non-invasive recordings & FM & PT + FT & 40,000+ hours electrical brain recordings & Yes \\
BrainWave~\cite{yuan2024brainwave} & FM for invasive and non-invasive recordings & FM & PT + FT & 13.79 TB electrical brain recordings (40,907 hours; 15,997 individuals) & Yes \\
BriVL~\cite{lu2022multimodal} & Multimodal FM bridging vision and language for neural encoding & FM & PT + FT & RUC-CAS-WenLan (30M image-text pairs), fMRI data & Yes \\
CBraMod~\cite{wang2024cbramod} & EEG decoding for BCI applications & FM & PT + FT & TUEG (1.1M samples, 9000+ hours) & No \\
Centaur~\cite{binz2024centaur} & Human behavior prediction & FM & PT + FT & Psych-101 (60K+ participants) & Yes \\
CodeBrain~\cite{ma2025codebrain} & EEG FM for interpretable decoding & FM & PT + FT & TUH EEG Corpus for pretraining and multiple EEG datasets & Yes \\
CSBrain~\cite{zhou2025csbrain} & Cross-scale spatiotemporal brain FM for EEG & FM & PT + FT & 11 EEG tasks across 16 datasets & Yes \\
DeepCNTD-Net~\cite{yoo2025non} & 3D FM for multi-label neuro-trauma detection on non-contrast head CT scans. & FM & PT + FT & Non-contrast head CT images & Yes \\
DECT~\cite{mo2025dect} & LLM-driven model extracting linguistic markers and synthesizing dialogue data & Framework w/ LLM & PT + FT & Speech recording and transcripts from ADReSSo & No \\
EEGFormer~\cite{chen2024eegformer} & Large-scale EEG FM for transferable learning & FM & PT + FT & TUH Corpus (1.7TB EEG dataset) & No \\
EEGPT (v1)~\cite{yue2024eegpt} & Generalist EEG FM for multiple tasks & FM & PT + FT & 37.5M pre-training samples, \textasciitilde1B tokens & Yes \\
EEGPT (v2)~\cite{yue2024eegpt} & Universal EEG feature extraction for medical and BCI & FM & PT + FT & Large mixed multi-task EEG dataset & Yes \\
EpiSemoLLM~\cite{yang2024episemollm} & Epileptogenic zone localization & LLM & PT + FT & Collected seizure semiology descriptions & Yes \\
ExKG-LLM~\cite{sarabadani2025exkg} & Cognitive neuroscience knowledge graph expansion & Framework w/ LLM & LLM-KG integration & Scientific papers and clinical reports & No \\
FM-HCT~\cite{zhu20253d} & 3D FM for head CT disease detection & FM & PT + FT & 361,663 non-contrast 3D head CT scans & Yes \\
FoME~\cite{shi2024fome} & FM for EEG with adaptive attention scaling & FM & PT + FT & 1.7TB diverse scalp and intracranial EEG & Yes \\
HybridTransNet~\cite{wu2024llm} & Brain tumor boundary delineation & FM & PT + FT & Medical imaging datasets & Yes \\
iReportMed~\cite{rashed2025automatic} & Hybrid pipeline: MRI features; LLM brain tumor diagnostic. & System w/ LLM & PT + FT & \textasciitilde378 brain tumor MRI scans (350 UCSF-PDGM + 28 clinical cases) & Yes \\
LaBraM~\cite{jiang2024large} & Large brain model for generic EEG representations & FM & PT + FT & \textasciitilde2,500 hours EEG from \textasciitilde20 datasets & Yes \\
LEAD~\cite{wang2025lead} & Clinical applications in neurodegenerative disease detection & FM & PT + FT & Alzheimer's disease related EEG datasets & Yes \\
LUNA~\cite{donerluna} & Topology-agnostic EEG processing with computational efficiency & FM & PT + FT & Diverse electrode configurations & Yes \\
Mental-LLM~\cite{xu2024mental} & LLM (GPT-based) fine-tuned to predict mental health status & System w/ LLM & PT + FT & Online text (social media data) & Yes \\
MindBridge~\cite{wang2024mindbridge} & Cross-subject brain decoding for visual reconstruction & FM & PT + FT & Multiple subjects' fMRI with visual stimuli & No \\
MindEye~\cite{scotti2023reconstructing} & fMRI-to-image reconstruction using contrastive learning & FM & PT + FT & Natural Scenes Image–fMRI Dataset (8 subjects) & No \\
MindEye2~\cite{scotti2024mindeye2} & Efficient fMRI-to-image with minimal training data & FM & PT + FT & Natural Scenes Image–fMRI Dataset (7 subjects) & No \\
MinD-Vis~\cite{chen2023seeing} & Visual stimulus decoding from fMRI using diffusion & FM & PT + FT & \textasciitilde136,000 fMRI samples, 1,205 subjects & Yes \\
MsLesionLLM~\cite{poole2025mslesionllm} & LLM-based prompt classifying MRI reports to extract new MS lesion information. & System w/ LLM & PT + FT & Clinical MRI report text & Yes \\
MultiViT~\cite{bi2024multimodal} & Structural-functional brain data integration for schizophrenia diagnosis & FM & PT + FT & Multi-site MRI (2,130 subjects) & Yes \\
MDD-LLM~\cite{sha2025mdd} & Major depressive disorder diagnosis & LLM & PT + FT & UK Biobank (274,348 records) & Yes \\
Neura~\cite{barrit2024neura} & Specialized neurology applications & LLM & PT + FT & Neurological corpus & No \\
Neuro-GPT~\cite{cui2023neuro} & FM for EEG-based BCI tasks & FM & PT + FT & TUH EEG Corpus, BCI Competition IV 2a & Yes \\
NeuroGPT-X~\cite{guo2023neurogpt} & Neurosurgical decision support & System w/ LLM & RAG-enhanced LLM & PubMed abstracts, Wikipedia & Yes \\
NeuroPictor~\cite{huo2024neuropictor} & fMRI-to-image reconstruction with semantic control & FM & PT + FT & 67,000 fMRI-image pairs & Yes \\
LaMIM~\cite{gao2025associations} & Postencephalitic epilepsy analysis using multi-contrast brain MRI & FM & PT + FT & 57,621 multi-contrast whole-brain MRI & Yes \\
PKG-LLM~\cite{sarabadani2025pkg} & Mental health prediction (GAD/MDD) & Framework w/ LLM & LLM-KG integration & NeuroLex, NeuroMorpho databases & Yes \\
POYO~\cite{azabou2023unified} & Spike-based neural decoding & FM & PT + FT & 158+ sessions, 27,373+ neural units & Yes \\
ProMind-LLM~\cite{zheng2025promind} & Psychiatric evaluation & System w/ LLM & PT + FT & PMData and Globem datasets & Yes \\
% scFoundation~\cite{hao2024large} & Single-cell transcriptomics representation learning & FM & PT + FT & Multiple single-cell RNA-seq datasets & Yes \\
SHADE-AD~\cite{fu2025shade} & LLM framework generating synthetic Alzheimer’s-specific human activity videos & Framework w/ LLM & Prompt-based & NTU RGB+D 120 dataset & Yes \\
SocialRecNet~\cite{chen2025socialrecnet} & Multimodal LLM (speech+text) framework predicting ASD social reciprocity scores & Framework w/ LLM & LLM integration & Speech and text for autism spectrum disorder & No \\
TRUST~\cite{tu2025trust} & LLM-powered dialogue system of cooperative modules for PTSD & System w/ LLM & LLM integration & Text (clinical interview dialogues) & No \\
VS-LLM~\cite{wu2024vs} & Vision-language LLM analyzing therapy sketches for depression assessment. & System w/ LLM & PT + FT & PPAT drawing test (sketches + stroke sequences) & No \\

\end{longtable}
}

\subsection{Brain-computer interfaces and neural decoding}
Brain computer interfaces (BCIs) are end-to-end systems that map neural measurements to control variables for an external device. The works discussed in this section study neural signal FMs and decoding frameworks that learn representations from large neural datasets, then adapt those representations to specific decoding tasks such as seizure detection and classification, short-term and long-term forecasting, or signal imputation. These models are not full BCI systems, and the available evidence is largely retrospective and benchmark based. As a result, they provide support for representation learning and transfer across datasets, while leaving open core BCI requirements such as reliable online adaptation under nonstationarity, calibration burden, and safety constraints in closed-loop use.

\textbf{Brain-computer interface models:} Self-supervised pretraining is frequently motivated by limited labeled data and by variability across subjects and recording sessions. BENDR is pretrained on the Temple University Hospital EEG Corpus using masked signal modeling with a contrastive predictive objective~\cite{kostas2021bendr}. The model adopts a two-stage architecture in which a stack of one-dimensional convolutional layers first compresses raw EEG into low-frequency representations at an effective sampling rate of approximately 2.67~Hz, followed by an eight-layer Transformer encoder that models long-range temporal dependencies in the compressed sequence. The pretraining corpus comprises recordings from more than 10{,}000 individuals with diverse clinical presentations and repeated sessions separated by intervals of up to eight months. In this work, the aggressive temporal downsampling biases the learned features toward slowly varying, context-dependent signal components, which are more stable across subjects and recording conditions than high-frequency activity. This design supports transfer to downstream tasks, including motor imagery decoding, P300 detection, error-related negativity classification, and sleep stage classification under standard fine-tuning protocols. At the same time, the observed gains primarily reflect improved representation quality within the evaluated benchmarks rather than evidence of modality-invariant generalization across all clinical EEG settings. In addition, the convolution-based positional encoding and long-context pretraining implicitly assume sequence lengths that exceed those typically encountered in short BCI trial paradigms.

CBraMod targets a similar goal of transferable EEG representations, but uses a different transformer backbone and a masked reconstruction objective. While both approaches target transferable EEG representations, BENDR employs a contrastive learning scheme with full-sequence modeling, whereas CBraMod uses a criss-cross transformer that models spatial and temporal dependencies separately through parallel attention mechanisms. This architectural difference allows CBraMod to capture heterogeneous spatial-temporal patterns more effectively while adapting to diverse channel configurations. CBraMod is pretrained on TUEG~\cite{wang2024cbramod}. The paper also emphasizes quality limitations in TUEG, including unmarked noise and artifacts, and describes substantial preprocessing that standardizes recordings to 19 channels, applies band-pass and notch filtering, resamples the data, and segments signals into fixed-length windows after excluding short and low-quality portions. Downstream evaluation spans multiple public datasets, although only a subset of results are shown in the main text. For emotion recognition, CBraMod reports balanced accuracy of $0.5509$ on FACED and $0.4091$ on SEED-V, alongside gains over the compared baselines under the paper's fixed experimental splits and multi-seed reporting. For motor imagery classification, the paper reports balanced accuracy of $0.6417$ on PhysioNet-MI and balanced accuracy of $0.6370$ with AUROC of $0.6988$ on SHU-MI. These results support improved cross-dataset decoding performance within the evaluated benchmarks. However, they do not, on their own, establish that the resulting models solve real-time BCI deployment problems, since online adaptation, latency constraints, and robustness to uncontrolled recording shifts are not directly tested in these experiments. 

FoME extends the foundation-model framing beyond classification to include short-horizon forecasting and imputation. The paper evaluates FoME on epilepsy-related classification tasks and reports, for seizure detection, accuracy of 95.16 with F2 score 95.15 on the MAYO dataset and accuracy 91.80 with F2 score 91.80 on FNUSA~\cite{shi2024fome}. For forecasting, the study evaluates short-term and longer-horizon prediction tasks and reports quantitative performance metrics including MAE and MSE values for both datasets, with FoME achieving, for example, short-term MAE 0.5675 and MSE 0.5874 on MAYO. For imputation, FoME reports substantial loss reductions relative to baselines, including MAE 0.1913 and MSE 0.1196 on MAYO. These experiments support the narrower claim that a single pretrained model can be adapted across multiple EEG tasks within controlled benchmark settings. They do not demonstrate clinical readiness for seizure forecasting or operational decision-making. The authors acknowledge limitations including dataset bias, under-representation of certain modalities, and open questions around broader forecasting capability in real-world clinical environments. In the context of BCI, FoME is best interpreted as a general-purpose neural time-series model that may inform decoding and signal modeling, rather than as evidence of deployable BCI systems.

Across these representative studies, clear trade-offs emerge in pretraining strategy and architectural design. Contrastive learning, such as BENDR, offers representation learning without explicit reconstruction but requires careful tuning of negative sampling and sequence length. Masked reconstruction, adopted in CBraMod and FoME, provides direct signal modeling but may overfit to dataset-specific noise patterns in TUEG. Both approaches achieve benchmark improvements over task-specific baselines, but the gains are task-dependent and modest in many cases. The central challenge is that retrospective benchmark performance does not predict robustness under conditions typical of real BCI deployment, including session-to-session drift, inter-subject variability in signal quality, and real-time latency constraints~\cite{patrick2025state}. Current evidence supports these models as tools for offline analysis and cross-dataset transfer, but not as validated solutions for closed-loop BCI applications.

\textbf{High-resolution neural decoding models:}
For high-channel, spike-resolved recordings, POYO proposes a scalable decoding framework that avoids assuming correspondence between neuron identities across sessions. The method tokenizes individual spikes and uses a PerceiverIO-style architecture with cross-attention to build latent representations that can be queried to regress behavioral variables~\cite{azabou2023unified}. The authors aggregate a multi-lab dataset spanning multiple cortical areas and recording sessions across nonhuman primates, and situate this scale as substantially larger than typical single-lab analyses. According to their evaluation protocol, the multi-session model achieves an average test $R^2$ of 0.9512 on center-out reaching and 0.8738 on random-target tasks. The paper also reports transfer results to held-out tasks and conditions, including $R^2=0.8962$ on NLB-Maze after unit identification, and $R^2=0.9329$ for a larger multi-task model variant using the same unit identification strategy. These results provide direct evidence that multi-session pretraining and lightweight session-specific adaptation can improve kinematic decoding performance across diverse spike datasets in nonhuman primates. However, they do not address several aspects that determine the practical performance of invasive BCIs, including long-term stability under chronic recording drift, human subject variability, and closed-loop robustness under real-time feedback and user learning. The contribution is therefore best understood as demonstrating scalable cross-session decoding methods for spike data analysis in nonhuman primates. The framework's applicability to chronic human BCI deployment remains to be established through validation under conditions of long-term recording drift, inter-subject variability, and closed-loop control. 

Across these neural decoding approaches, two persistent challenges limit translation from benchmark performance to deployed BCI systems. First, the evaluation-deployment gap emerges when models are tested on curated retrospective data with known labels and fixed recording conditions, but must operate prospectively under signal drift, user adaptation, and variable environmental interference. Second, the cross-species transfer problem appears when models trained or validated on nonhuman primate data are expected to generalize to human subjects, who exhibit different cortical organization, task strategies, and recording stability profiles. These challenges are compounded by the fact that benchmark metrics (accuracy, $R^2$, MAE) do not capture user experience, learning curves, or long-term usability, which are critical for clinical BCI adoption~\cite{patrick2025state}. Future work may benefit from prospective validation protocols that assess performance under realistic deployment conditions, including extended recording sessions, user training periods, and integration with assistive devices.

\subsection{Clinical decision support and evidence interfaces}
A growing subset of foundation-model work in neuroscience targets clinical knowledge use, documentation interpretation, and evidence-linked explanation generation, rather than neural signal modeling. These efforts typically instantiate a \emph{system} that couples a language model with domain curation, retrieval, and citation surfacing. Their empirical support comes largely from retrospective case collections, exam-style question banks, or vignette-based evaluations, and rarely extends to prospective validation, workflow integration studies, or patient outcome analyses. The absence of such validation reflects broader challenges in translating AI tools into clinical practice, including distribution shifts across clinical sites, incomplete documentation practices, and the gap between retrospective accuracy and prospective safety~\cite{dinsdale2022challenges,leming2023challenges}. The central technical question is not whether a model can generate fluent text, but whether it can map clinical narratives into clinically meaningful structured hypotheses and sustain evidence-grounded reasoning under domain shift, incomplete context, and adversarial inputs. Reported benchmarks often emphasize accuracy or rubric-based ratings, while robustness indicators, calibration, and failure characterization remain unevenly reported. Where comparisons exist, they are typically confined to narrowly scoped tasks and small, curated test sets, so performance differences should be interpreted as task- and protocol-specific rather than as general clinical readiness.

\subsubsection{Clinical decision support systems}
One representative design pattern is to treat clinical narratives as inputs to structured hypothesis generation. EpiSemoLLM frames seizure semiology interpretation as a text-to-localization prediction task and was fine-tuned from Mistral-7B on 865 literature-curated semiology–EZ pairs with surgically validated labels~\cite{yang2024episemollm}. In a 100-case retrospective comparison with five epileptologists, the model showed region-dependent performance, exceeding the human average in temporal and insular localization but lagging in frontal, occipital, and cingulate regions. The evaluation remains limited by its rubric-based design, small expert cohort, and lack of testing under heterogeneous clinical documentation and missing data.

A second pattern emphasizes subspecialty question answering with an explicit evidence interface. NeuroGPT-X illustrates a context-enriched chat system for vestibular schwannoma management. The study reports web-based curation comprising 157 Wikipedia articles, divided into 1,659 sections, and literature retrieval that returned 4,004 PubMed publications, along with smaller numbers from other sources~\cite{guo2023neurogpt}. A senior neurosurgeon curated 15 questions, four neurosurgeons and two GPT variants generated responses, and three independent neurosurgeons rated answers on 0 to 4 Likert scales for accuracy, coherence, relevance, thoroughness, and overall quality. The authors report that naive and context-enriched GPT responses were often rated higher than expert responses on these rubric scores, and that model responses were faster than expert respondents (p<0.01). They also report agreement with 98 of 103 consensus statements (95\%) from a set of statements on vestibular schwannoma. However, all expert surgeons expressed concerns about the model's reliability in addressing nuances and controversies in the management of vestibular schwannoma. This evidence supports the feasibility of subspecialty, citation-facing information synthesis under a narrow domain, but a 15-question set and Likert ratings do not quantify clinical error modes or safety failures, and consensus-statement agreement does not demonstrate safe handling of patient-specific contraindications.

A more constrained, exam-style setup provides a different view of capability and robustness. AtlasGPT provides an evaluation protocol closer to exam-style assessment than open-ended consult-style dialogue. In a 149-question neurosurgery examination, AtlasGPT reported 96\% accuracy, compared with 93\% for Gemini Advanced and 88\% for GPT-4~\cite{hopkins2024atlasgpt}. Under an adversarial misinformation-identification setup, AtlasGPT was reported to be fooled 14\% of the time, compared with 44\% for GPT-4 and 68\% for Gemini Advanced. Explanations were rated by 15 neurosurgeons as more comprehensive and better referenced than the question bank explanations (p<0.001). This evidence supports robustness advantages within the specific adversarial protocol used, but the endpoint remains question-answer correctness and explanation quality rather than prospective clinical decision impact. The exam-style format does not capture the ambiguity, incomplete information, and time pressure characteristic of real clinical encounters.

Retrieval-augmented systems extend this interface logic beyond a single subspecialty and into multi-domain vignette testing. Almanac demonstrates how retrieval and structured evidence presentation can change the evidentiary profile of a general medical question-answering system. Using a novel dataset of 130 clinical questions spanning five medical specialties (ClinicalQA) and evaluations by 15 internists (mean 14 years in practice), Almanac was reported to increase factuality by 18 percentage points relative to ChatGPT (p=0.0005), with the largest reported gain in cardiology (91\% vs 69\%, p=0.018856)~\cite{zakka2024almanac}. Completeness improved by 4.8 percentage points but was not statistically significant (p=0.25). In a safety-oriented adversarial setting, Almanac responses were rated safe 95\% of the time, compared with 0\% for ChatGPT (p<0.001), and in five clinical calculation scenarios Almanac was reported correct in all five while ChatGPT was incorrect in all five. The same study reports that ChatGPT supported its statements with correct sources only 56\% of the time, yet clinicians still preferred ChatGPT answers 57\% of the time. This preference-factuality gap indicates that retrieval can strengthen factual grounding under a defined protocol, while user preference remains influenced by factors beyond factual accuracy alone.

These systems illustrate distinct technical trade-offs in clinical decision support design. Fine-tuning offers domain specialization but requires curated training data and may not generalize beyond the training distribution. EpiSemoLLM, for instance, shows heterogeneous regional performance that likely reflects data scarcity in rare anatomical sites like the cingulate. Retrieval-augmented generation preserves broader knowledge and enables citation surfacing, but introduces latency, citation mismatch, and reliance on curated corpora, as evidenced by systems like NeuroGPT-X, AtlasGPT, and Almanac. Exam-style evaluation provides reproducible benchmarks but fails to capture clinical ambiguity, while vignette-based approaches better approximate decision complexity but remain vulnerable to gaps between preferences and facts. The consistent pattern across these approaches is that systems optimized for factual accuracy do not automatically gain clinical acceptance, suggesting that trust, workflow integration, and explanation quality may be orthogonal to technical correctness. This points to fundamental gaps in how clinical utility is currently operationalized and measured.

\subsubsection{Evaluation benchmarks and knowledge-graph frameworks}
Other work focuses on knowledge infrastructure rather than interactive support. ExKG-LLM proposes an automated knowledge-graph extraction pipeline and reports extraction performance at precision 0.80, recall 0.81, and F1 0.805 on the stated benchmark, alongside graph-structural improvements after optimization~\cite{sarabadani2025exkg}. These metrics quantify information extraction quality, but they do not by themselves validate clinical correctness of the underlying statements or patient-level utility. Their practical value depends on how extracted relations are curated, updated, and audited in downstream use.

Verification-oriented neuro-symbolic refinement offers a different kind of control than retrieval alone. PEIRCE uses neuro-symbolic refinement with theorem proving to iteratively verify and improve explanations. On premise-selection style evaluation with 50 randomly selected examples from WorldTree and ProofWiki, the reported mean average precision for an ensemble retriever exceeded BM25 (WorldTree: 38.72 vs 22.84 and ProofWiki: 27.09 vs 10.18)~\cite{quan2025peirce}. The same work describes iterative verification experiments where the number of verified explanations increases over iterations, including in a clinical explanation setting~\cite{quan2025peirce}. This approach improves verifiability under a formal proof system for tasks that can be expressed with suitable premises and logic forms. It does not establish that the verified explanations correspond to clinically complete reasoning, nor does it address missing-context failures common in real documentation. The formalization requirement may exclude clinically relevant reasoning patterns that resist logical decomposition.

Mental-health oriented pipelines illustrate how extraction and graph completion can be coupled to downstream discriminative tasks. PKG-LLM extends knowledge-graph construction to mental health-oriented clinical text. This work reports 430 documents from 178 patients, with entity and relation extraction F1 scores of 0.93 and 0.90, and link prediction AUC 0.92 using a graph-embedding component~\cite{sarabadani2025pkg}. These results support structured extraction and graph completion on the reported dataset, but they remain an intermediate representation result. The distinction between link prediction in a knowledge graph (inferring missing edges) and clinical prediction (forecasting patient outcomes or diagnostic categories) is critical. These extraction metrics should not be conflated with disorder classification accuracy under external validation or with clinical benefit.

A separate strand aims at cognitive and behavioral modeling rather than diagnosis or documentation support. Centaur is trained on Psych-101, described as trial-by-trial data from 160 psychological experiments involving 60,092 participants and 10,681,650 choices, and it reports improved generalization across task variants and new domains relative to prior cognitive models~\cite{binz2024centaur}. This is a model of behavioral response distributions, not a neural signal model. The evidence supports cognitive modeling claims rather than mechanistic inference about brain dynamics or clinically validated assessment. Cognitive models capture statistical regularities in behavior but do not necessarily identify underlying neural mechanisms, nor do they provide actionable clinical predictions without further validation in clinical populations.

Finally, benchmarking efforts aim to standardize what it means for models to perform neuroscientific reasoning. Benchmark proposals such as BrainBench aim to operationalize evaluation for neuroscience- and medicine-adjacent reasoning~\cite{luo2024large}. Without consistent reporting of task composition, gold-standard construction, and safety constraints across benchmarks, cross-paper comparisons remain fragile, and benchmark scores should be treated as protocol-dependent indicators rather than transferable evidence of clinical competence. The proliferation of benchmarks without standardized evaluation protocols or clear clinical grounding creates a risk of optimization for benchmark performance without corresponding improvements in real-world utility or safety~\cite{huang2025ai}. 

While these systems show impressive performance on curated benchmarks, their translation to clinical practice faces recurring obstacles. Three recurring patterns characterize the current deployment challenges in this domain. First, evaluation-practice mismatch occurs when retrospective accuracy metrics are trained and tested on curated, complete documentation, but real clinical encounters involve missing information, ambiguous phrasing, and contradictory cues systematically excluded from current benchmarks. Second, the preference-factuality paradox emerges when clinicians prefer ChatGPT (57\%) despite Almanac's superior factuality (91\% vs 69\%), suggesting that fluency and cognitive fit dominate trust formation more than verifiable correctness. Third, domain generalization failure occurs when performance heterogeneity across anatomical sites (EpiSemoLLM's 83\% temporal vs. 28\% cingulate accuracy) reflects not merely data scarcity but also fundamental challenges in capturing rare, high-stakes scenarios where subtle cues and anatomical complexity intersect. These persistent tensions between retrospective accuracy and prospective safety, between factual correctness and clinical acceptance, and between benchmark performance and rare-case robustness point to fundamental gaps in how clinical AI utility is currently conceptualized and measured~\cite{dinsdale2022challenges,huang2025ai}. These benchmark and infrastructure directions provide a bridge to downstream sections that focus on disease-specific modeling and application contexts, where evidence strength is typically determined by dataset design, external validation, and clinical endpoints.

\begin{figure}
\centering
\includegraphics[width=0.99\linewidth]{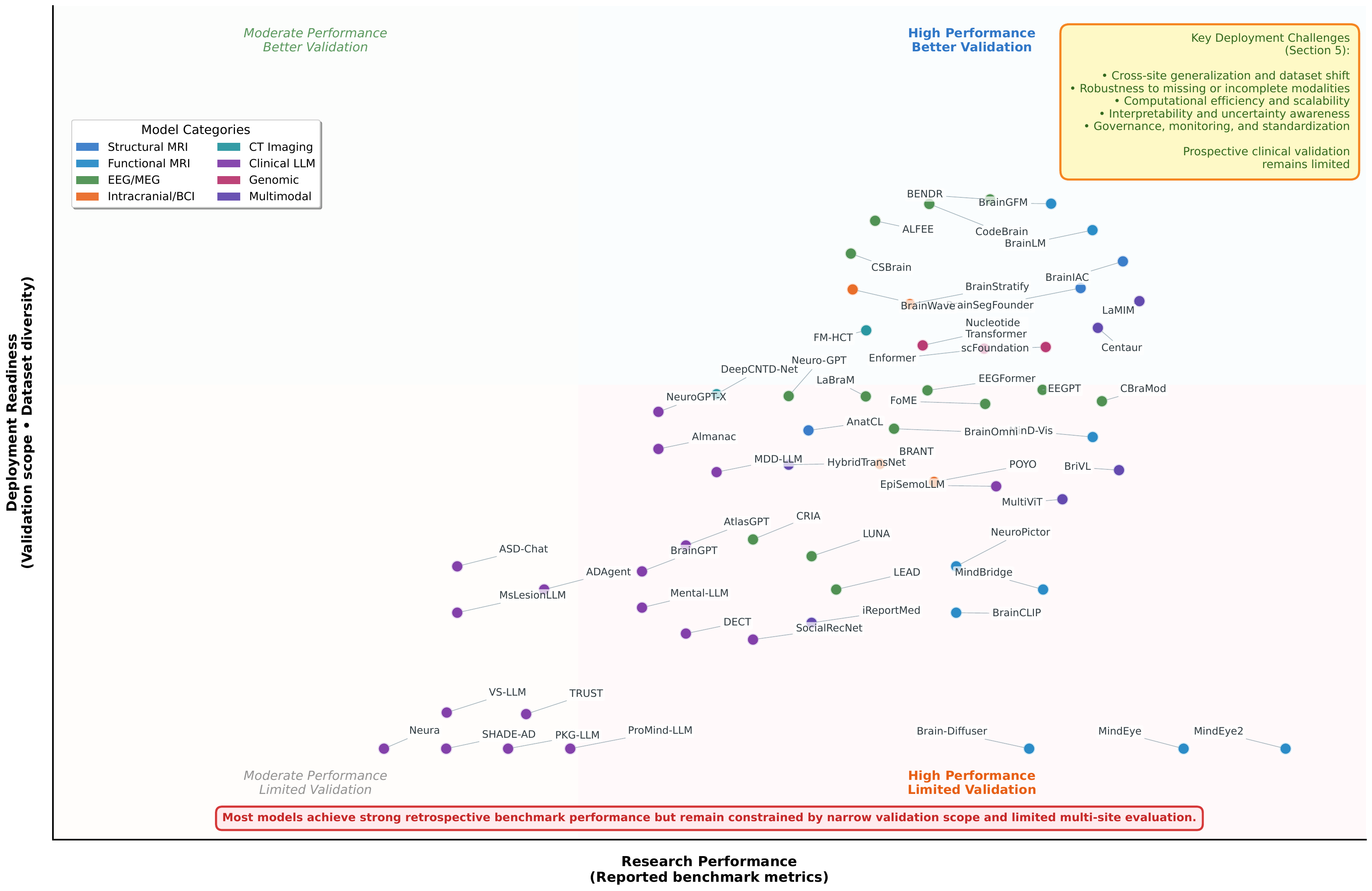}
\captionof{figure}{Schematic landscape of neuroscience AI models: Research Performance vs. Deployment Readiness. Each point represents a reviewed model, color-coded by modality. Positioning Methodology: The landscape utilizes a semi-quantitative binning strategy to visualize developmental stages. The horizontal axis (Research Performance) reflects the qualitative improvement reported over domain-specific state-of-the-art baselines. The vertical axis (Deployment Readiness) denotes the scope of validation evidence as summarized in Table 4, progressing from internal retrospective evaluation to multi-center validation and prospective clinical deployment. Because evaluation metrics and task difficulty vary substantially across studies, relative distances between models should be interpreted as conceptual differences in developmental maturity rather than as precise quantitative comparisons.}
\label{fig:eval_gap}
\end{figure}

\subsection{Disease-specific applications}
Disease-specific neuroscience research is increasingly exploring large-scale AI models for brain disorder diagnosis and treatment, both as computational tools for focused tasks and as elements within broader clinical systems. Early evidence remains preliminary. For instance, Luo et al. examined the use of ChatGPT in presurgical decision-making for epilepsy to evaluate whether LLMs could assist in interpreting complex seizure semiology within clinical workflows~\cite{luo2025clinical}. Related efforts include applying pretrained models to disease classification, fine-tuning architectures for condition-specific objectives, and incorporating language models into clinical decision-support prototypes. LLMs have also demonstrated competence in standardized settings, with ChatGPT-4 achieving passing scores on neurology examinations and performing well on cognitive assessments~\cite{schubert2023performance,moura2024implications}. In the BrainBench benchmark, these models were able to predict outcomes of neuroscience experiments, although the extent to which this capability translates into real scientific discovery remains uncertain~\cite{luo2024large}. Despite these encouraging results, their clinical effectiveness, safety, and impact on patient outcomes in neurological and psychiatric disorders remain to be rigorously validated.

In practice, disease-oriented systems rarely deploy FMs as standalone diagnostic tools. Instead, they are embedded in task-specific pipelines that combine transferable representations or structured language priors with disease-aware components that encode pathological signatures, temporal progression, and modality-driven biomarkers. System performance therefore reflects not only model scale but the extent to which general representations are aligned with the biological characteristics and clinical objectives of each disorder.

\subsubsection{Neurodegenerative disorders}

Neurodegenerative conditions pose challenges for AI-assisted diagnosis and prognosis due to their progressive nature and complex pathophysiology. While AI approaches are being developed for Alzheimer's disease, Parkinson's disease, and multiple sclerosis using disease-specific biomarkers and progression patterns, their clinical utility remains under investigation.

While AI methods have achieved notable performance in discrete classification tasks such as early detection, diagnosis (e.g., AD vs. non-AD), and predicting conversion to dementia~\cite{bron2022ten, ansart2021predicting}, their application to continuous and fine-grained prognostic measures remains limited. Predicting gradual trajectories of neurological and cognitive decline requires models that capture subtle temporal and physiological variations, rather than merely performing binary classification. Current approaches show only modest success. For instance, neuroimaging-based models often struggle to accurately predict longitudinal changes in cognitive scores~\cite{SUI2020818}. Similarly, transfer learning based on a pretrained ResNet50 model achieved reliable conversion prediction but exhibited weak correlations with actual cognitive decline~\cite{bae2021transfer}. Addressing these challenges is crucial for advancing AI's role in clinical settings and for deepening our understanding of fundamental questions in the neuroscience of dementia~\cite{thivierge2007topographic}, such as how neural signal features map onto higher-order cognitive processes.

ADAgent represents an emerging class of LLM-driven conversational systems designed to support Alzheimer’s Disease analysis through multi-turn interaction and tool orchestration~\cite{hou2025adagentllmagentalzheimers}. Rather than operating as a standalone diagnostic model, ADAgent coordinates a set of specialized multimodal tools through reasoning, planning, and a collaborative outcome coordinator, enabling flexible diagnosis and prognosis under complete or missing modality settings. By adapting its interaction strategy to different stages of cognitive impairment, the system aims to support early-stage monitoring and clinician-facing decision support. At a more model-centric level, DECT demonstrates how LLMs can enhance speech-based Alzheimer’s Disease detection through fine-grained linguistic analysis and data augmentation~\cite{mo2025dect}. DECT leverages LLMs to distill cognitive-linguistic markers from noisy transcripts and to generate label-switched or label-preserved synthetic speech data, which are then used to improve transformer-based classification performance. From a representation-learning perspective, this approach enriches task-specific linguistic features rather than replacing conventional classifiers. It also illustrates complementary roles for LLMs in AD research, spanning system-level reasoning and coordination on one hand, and targeted representation and data enhancement for specific diagnostic tasks on the other.

Also of notable relevance to Alzheimer’s Disease research is the Synthesizing Human Activity Datasets Embedded with AD Features (SHADE-AD) framework, which addresses the lack of disease-specific behavioral data by generating synthetic human activity datasets enriched with Alzheimer characteristic motor patterns~\cite{fu2025shade}. SHADE-AD adopts a three-stage LLM-assisted training pipeline that progressively integrates generic action semantics, domain-specific AD knowledge, and fine-grained kinematic constraints. In the first stage, large-scale human activity corpora are used to establish general motion primitives. The second stage injects Alzheimer related behavioral descriptors through LLM-guided supervision, biasing the generation process toward subtle deviations associated with cognitive and motor decline. The final stage enforces joint-level motion consistency using quantitative kinematic metrics, ensuring physical plausibility and temporal coherence of synthesized movements. SHADE-AD demonstrates how LLMs can serve as structured knowledge translators rather than end-to-end predictors, embedding clinical domain knowledge into data-generation pipelines in a controllable manner. By operating at the level of behavior synthesis rather than diagnosis, the framework enables downstream human activity recognition models to better capture AD-relevant motion patterns in smart health monitoring scenarios. At the same time, the scope of applicability remains limited to activity-level inference, and the relationship between synthesized behavioral features and underlying neuropathology is not directly modeled.

Providing a broader foundation for this line of research, Gao et al. present a comprehensive review of large language model applications across neurodegenerative disorders, including Alzheimer’s disease, Parkinson’s disease, and multiple sclerosis~\cite{gao2024llms}. Rather than proposing a new model, the study contributes a systematic analysis of how LLM-based systems are evaluated, benchmarked, and interpreted across heterogeneous clinical tasks. The review highlights recurring challenges related to data heterogeneity, limited reproducibility across datasets, interpretability of model outputs in clinical settings, and regulatory considerations for deployment. This analysis clarifies that many reported performance gains are tightly coupled to task-specific benchmarks and annotation protocols, underscoring the need for standardized evaluation frameworks that go beyond accuracy metrics. By explicitly framing these issues, the work provides methodological guidance for future LLM-based systems in neurodegenerative disease research, emphasizing that progress depends not only on model scale but also on principled validation and clinical alignment.

In the context of multiple sclerosis, MsLesionLLM exemplifies the use of language models for structured extraction of disease activity from unstructured clinical documentation~\cite{poole2025mslesionllm}. The system applies an LLM-based prompt to radiology reports to identify new T2-weighted and contrast-enhancing lesions, enabling longitudinal monitoring of inflammatory activity and treatment response in real-world datasets. While the reported performance demonstrates high accuracy in retrospective validation, the approach operates at the level of report interpretation rather than direct image analysis, and its clinical utility depends on consistency in reporting practices and prospective evaluation.

\subsubsection{Neurological and cerebrovascular disorders}
Neurological and cerebrovascular disorders that arise from external trauma, vascular events, or pathologic processes require timely diagnosis and treatment, making them potential areas for AI-assisted clinical decision support. These disorders often present complex symptom patterns that researchers are exploring using AI-based pattern recognition and data analysis, though clinical validation remains ongoing.

Epilepsy, as a heterogeneous neurological disorder involving multimodal data streams spanning clinical records, EEG, video, neuroimaging, and genomics, has motivated growing interest in AI-based approaches for detection, diagnosis, localization, risk stratification, and treatment planning. Despite this breadth of investigation, the clinical effectiveness of such systems across the epilepsy care pathway remains incompletely validated. Video-based seizure detection has emerged as a complementary sensing modality, particularly in settings where long-term EEG monitoring is impractical due to cost, portability, or patient burden. SETR-PKD introduces a transformer-based framework that operates on optical flow representations derived from RGB video, explicitly framing motion cues as privacy-preserving surrogates for seizure-related behavior~\cite{mehta2023privacy}. A progressive knowledge distillation strategy transfers temporal representations from models trained on longer seizure segments to those operating on shorter segments, enabling earlier detection without relying on full seizure recordings. Within the specific task of tonic-clonic seizure detection, the method achieves an accuracy of 83.9\% when evaluated halfway through seizure progression. From a modeling perspective, the use of optical flow constrains the representation to gross motion dynamics while discarding appearance cues, thereby protecting privacy and limiting sensitivity to fine-grained motor patterns. As a result, the reported performance supports feasibility for early detection of a narrowly defined seizure type under controlled conditions, but does not establish robustness across heterogeneous home environments, camera viewpoints, or seizure semiologies with subtler motor expression.

VSViG explores a different representational trade-off by formulating video-based seizure detection as a skeleton-centric spatiotemporal graph learning problem~\cite{xu2024vsvig}. By embedding joint-centered patches and modeling their temporal evolution with a Vision Graph architecture, the approach prioritizes efficiency and low-latency inference while retaining sensitivity to structured body movements. Reported gains over prior baselines are accompanied by reductions in parameter count and computational cost, highlighting suitability for real-time deployment scenarios. However, as with optical-flow-based methods, the learned representations remain fundamentally behavioral rather than neurophysiological. Consequently, video-based detection should be interpreted as a pragmatic sensing strategy that can support seizure monitoring in constrained settings, rather than as a substitute for EEG when clinical questions require characterization of underlying brain dynamics.

For diagnostic characterization, the current evidence remains limited and highly system-dependent. EpilepsyLLM is presented as a disease-specific large language model fine-tuned on epilepsy-related medical knowledge, to improve the reliability and relevance of textual responses in this domain~\cite{zhao2024epilepsyllm}. Rather than operating as a multimodal diagnostic system, the model focuses on language-based understanding and generation, drawing on curated epilepsy knowledge covering disease background, common treatments, and practical considerations for daily life. Experimental evaluation is conducted using standard natural language generation metrics, demonstrating improved domain-specific response quality compared with general-purpose and medical LLM baselines. The current evidence base for EpilepsyLLM is limited by its evaluation scope, which focuses on language-generation benchmarks rather than clinical outcome studies or prospective validation. As a result, claims regarding treatment optimization or therapeutic planning should be interpreted with caution.

In early epilepsy evaluation, seizure semiology provides clinically useful information for presurgical localization~\cite{elwan2018lateralizing}. EpiSemoLLM is a fine-tuned language model that infers likely epileptogenic zone from seizure semiology narratives and limited structured context~\cite{yang2024episemollm}. Built on Mistral-7B-Instruct and fine-tuned with LoRA on 765 literature-derived semiology--EZ pairs, the model is evaluated on a 100-case hold-out set using task-specific metrics, including the rectified Reliability Score (rRS) and Regional Accuracy Rate (RAR). Under zero-shot prompting, it reports a mean rRS of 0.291, exceeding the average rRS of board-certified epileptologists in the same evaluation, with the highest RAR in temporal lobe cases and markedly lower accuracy in rare frequent regions such as the insular and cingulate cortex. Even though the model works well on its evaluation scenario, it does not establish physiological localization, sublobar precision, or prospective clinical utility.

Risk modeling in at-risk populations provides a clearer illustration of how FMs can be applied to retrospective classification tasks under explicit evaluation protocols. Gao et al.\ develop a large self-supervised vision FM using 57{,}621 multi-contrast head MRI scans from 34{,}871 patients for representation learning, followed by fine-tuning on a cohort of 144 encephalitis patients, including 64 with postencephalitic epilepsy and 80 without, with a median follow-up of 3.7 years (range 0.7--7.5)~\cite{gao2025associations}. In this retrospective setting, the association model achieves an accuracy of 79.3\%, sensitivity of 92.3\%, specificity of 68.8\%, F1 score of 80.0\%, and an AUC of 81.0\%, with a statistically significant improvement over a DenseNet model trained from scratch (DeLong test $P=0.03$). The results support the premise that large-scale self-supervised pretraining on heterogeneous, unlabeled MRI data can mitigate the limitations imposed by small labeled datasets, particularly for diseases characterized by diffuse and spatially variable imaging abnormalities. At the same time, the evidence remains confined to retrospective association within a single study population. The reported performance does not establish prospective risk-stratification utility, calibration for individual-level decision-making, or robustness across institutions and imaging protocols beyond the evaluated datasets.

Treatment selection remains a separate translational problem where evidence of real-world benefit is still emerging. A protocol described by Thom et al. outlines a multisite, prospective, double-blind randomized controlled trial evaluating a machine-learning decision-support tool for first antiseizure medication selection, targeting recruitment of at least 234 adults across 14 centers in Australia~\cite{thom2025personalised}. As a protocol, this work does not yet provide outcome evidence and should be cited as an ongoing test of clinical utility rather than proof of effectiveness. In parallel, retrospective modeling studies suggest that integrating genetic with clinical information can improve discrimination for seizure-freedom outcomes compared with clinical-only baselines. Feng et al. report, in a development cohort of 286 participants, an AUC of 0.74 for a clinical+genomic model versus 0.67 for clinical-only features, and in an external validation cohort of 219 participants, an AUC of 0.69 versus 0.62~\cite{Feng2025Integrative}. This evidence supports incremental predictive gains under retrospective evaluation, but it does not resolve challenges central to deployment, including dataset shift across healthcare systems, missingness and bias in genetic testing, and the need to demonstrate that model-guided prescribing improves patient-centered outcomes beyond standard care.

Related work in acute stroke and broader neurology primarily focuses on early-stage decision-support systems and analytical workflows rather than clinically validated, deployable tools. Song et al.\ present an LLM-centered framework that integrates clinical presentation with neuroimaging reports to support time-critical assessment and reasoning in acute stroke care~\cite{song2025stroke}. The system is evaluated retrospectively using internal and cross-institutional datasets, demonstrating feasibility for stroke identification, subtype classification, and treatment-related decision support. However, the evaluation remains limited to offline validation, and the model is explicitly framed as an auxiliary diagnostic tool rather than an autonomous clinical system. Kottlors et al. similarly investigate the use of large language models for report-based thrombectomy decision-making and outline potential deployment patterns across neurological workflows~\cite{kottlors2025large}. Their study emphasizes feasibility and human-in-the-loop support, highlighting the role of LLMs as background monitoring or alerting systems rather than primary decision-makers. Taken together, these contributions are best interpreted as systems-oriented explorations of support for clinical reasoning. In the absence of prospective trials, regulatory review, or demonstrated impact on real-world workflow efficiency and patient outcomes, they should be framed as decision-support concepts and hypothesis-generation interfaces rather than near-term clinically deployable solutions.

\subsubsection{Brain tumors}

Surgery planning and brain tumor diagnosis are areas where AI applications are being investigated for their potential to assist with diagnostic assessment, treatment planning, and surgical navigation. AI systems are being developed to analyze neuroimaging data and integrate with clinical reporting systems, though the clinical impact on patient outcomes requires further validation. Recent FM approaches for brain tumor assessment span two complementary methodological directions. One line of work focuses on image-centric representation learning for tasks such as tumor classification, retrieval, and segmentation, while a parallel line investigates the use of language models to support interpretation, reporting, and clinician interaction rather than direct multimodal data fusion. Manjunath et al.\ propose an interactive image retrieval framework for brain tumor classification that combines supervised contrastive image representations with large language model-based summarization and question answering over retrieved cases~\cite{manjunath2024towards}. The approach remains primarily image-driven, with LLMs serving as an interpretability and accessibility layer rather than as a mechanism for fusing heterogeneous biomedical modalities. From a methodological perspective, the work emphasizes preservation of latent structural similarity in the retrieval space to support example-based reasoning, but does not establish prognostic inference or outcome prediction beyond retrospective classification benchmarks. In parallel, HybridTransNet introduces a hybrid convolutional and transformer-based architecture for multi-sequence MRI analysis, framing boundary delineation and volumetric quantification as central objectives~\cite{wu2024llm}. By integrating local feature extraction with global attention, the model aims to improve segmentation performance across heterogeneous tumor appearances. However, reported accuracy and uncertainty characteristics are inherently tied to the specific datasets and evaluation protocols used, and should not be interpreted as a general guarantee of clinical reliability without broader validation.

Language-centric systems, by contrast, primarily model clinical narratives and their internal consistency rather than tumor biology directly. BrainGPT exemplifies this direction as a vision-language report generation framework for volumetric brain CT captioning and structured radiology-style descriptions, evaluated using the authors’ feature-oriented radiology task evaluation (FORTE) metric, together with judge-based and physician assessments~\cite{li2025towards}. In the reported experiments, training uses 365{,}928 slices sampled from 15{,}238 scans (7{,}747 patients) and testing uses 87{,}312 slices from 3{,}638 scans (1{,}938 patients), with an external zero-shot evaluation on CQ500 restricted to non-contrast scans with 23--40 slices ($n=133$). On CQ500, the instruction-tuned variants achieve keyword-level reporting accuracy of 0.71–0.75 for mass effect, 0.35–0.38 for midline shift, and 0.43–0.50 for hemorrhagic events. Applying a negation-removal preprocessing step further increases the accuracy to 0.86–0.91 for midline shift and 0.59–0.61 for hemorrhagic events. The study explicitly positions these results as evidence of report-generation fidelity rather than diagnostic performance and presents them as a pilot without comparison to established multimodal large language models. Moreover, the degeneration-oriented training distribution constrains coverage of malignant tumors and acute traumatic findings in CQ500. Therefore, the results support the narrower conclusion that vision-language models can recover elements of radiology style linguistic structure and selected abnormality descriptors under controlled evaluation, while leaving open questions regarding robustness, generalization, and clinical decision-making utility.

A related but distinct use of language modeling is to treat reports as structured evidence sources for downstream classification and survival stratification. Ma et al. develop an LLM-based multi-source integration pipeline that extracts features from radiology and pathology reports and trains an MLP for supervised classification of tumor presence and tumor stability on a cohort of 426 patients, reporting micro-F1 of 0.849 for tumor presence and 0.929 for tumor stability, with AUROC 0.893 and 0.964 respectively~\cite{ma2025large}. The same study evaluates a zero-shot prognosis setting on an independent cohort of 33 glioblastoma patients by stratifying risk using model-produced label probabilities and comparing against MGMT promoter methylation and radiomics-derived factors, reporting a significant survival separation when using the predicted tumor stability label (log-rank p=0.017). The evidence therefore supports the feasibility of text-derived multimodal report integration for label-based stratification under the stated datasets and assumptions, while leaving open how robust these signals remain under broader multi-institutional reporting heterogeneity or prospective workflows. Besides, iReportMed targets automatic diagnostic report generation from multimodality MRI through a two-stage pipeline that combines multi-task image analysis with LLM-conditioned text generation, using the UCSF-PDGM dataset (542 subjects; a 374-subject glioblastoma-confirmed subset is described) and a smaller clinical dataset (28 subjects) for fine-tuning and evaluation~\cite{rashed2025automatic}. The report-generation model is trained in two phases using 350 UCSF-PDGM subjects and then fine-tuned on 18 clinical subjects, with evaluation on the remaining 10 clinical subjects by two external radiologists scoring completeness, correctness, and conciseness on a 0--10 scale. The paper also reports an ablation focused on linguistic quality, where adding an image perception model improves METEOR by 7.2\% and C-BLEU by 2.8\%. At the same time, the authors explicitly note practical limitations including variability across MRI acquisition protocols and radiologist reporting styles, and recommend human-in-the-loop verification to mitigate hallucination risk, which constrains any clinical readiness claims to decision-support and documentation-assistance settings rather than autonomous reporting.

\subsubsection{Neuropsychiatric disorders}
Understanding neuropsychiatric disorders requires examining patient behavior, symptoms, and treatment responses beyond just measuring biomarkers. AI models are being developed and tested to analyze natural language, behavioral data, and health records, with the goal of potentially improving diagnostic assessment and treatment planning. However, clinical validation and real-world implementation of these approaches remain limited.

Targeting major depressive disorder specifically, MDD-LLM formulates depression assessment as a supervised classification task over structured health records by transforming UK Biobank tabular variables into natural-language prompts for parameter-efficient fine-tuning. Using 274{,}348 individual records, the 70B model reports an accuracy of 0.8378 and an AUC of 0.8919, outperforming conventional machine learning and deep learning baselines evaluated on the same dataset~\cite{sha2025mdd}. Methodologically, this design leverages the linguistic prior of large language models to integrate heterogeneous demographic, behavioral, and biochemical features without explicit feature engineering, while retaining probabilistic outputs that support risk stratification. At the same time, model performance remains tightly coupled to the dataset definition and labeling strategy derived from retrospective ICD-10 codes and self-reported measures. The reported gains reflect improved feature utilization within a fixed observational dataset rather than validation of real-world diagnostic equivalence or prospective clinical utility.

A complementary line of work explores drawing-based projective tests as an indirect probe of depressive state. VS-LLM operationalizes the Drawing Projection Test using PPAT sketches by first extracting visual cues related to stroke dynamics, color usage, and spatial layout, and then prompting a large language model to generate psychologically oriented semantic captions that are fused with visual features for depression versus non-depression classification~\cite{wu2024vs}. On a PPAT dataset comprising 690 sketches, the method achieves 87.8\% accuracy, a 17.6 percentage-point improvement over the strongest psychologist-score-based baseline (70.2\% accuracy). This approach leverages language models to encode expert-informed semantic abstractions that are difficult to specify as fixed visual features, thereby complementing low-level stroke and layout representations. However, evaluation is confined to a single dataset and experimental protocol, with no evidence that performance generalizes across sites, raters, or variations in drawing conditions.

Beyond disorder-specific models, Mental-LLM evaluates and instruction-tunes LLMs for multiple mental health classification tasks on online textual data, demonstrating that instruction fine-tuning improves balanced accuracy relative to zero-shot and few-shot prompting baselines~\cite{xu2024mental}. The reported gains are substantial in aggregate comparisons, but the tasks remain defined as retrospective text classification benchmarks with fixed labels and predefined task boundaries. The authors explicitly note that these results do not imply deployability, highlighting concerns related to bias, generalization, and ethical risk. As a result, the primary contribution lies in characterizing how instruction tuning alters benchmark performance across tasks, rather than establishing reliability for real-world mental health assessment. ProMind-LLM extends this benchmark-oriented line of work by incorporating passive sensor data alongside self-reported mental records and introducing causal-style reasoning prompts for binary mental health risk assessment~\cite{zheng2025promind}. The model is evaluated on two open datasets with expert-reviewed label construction, including PMData with 16 participants over five months and Globem with 497 participants over four years. While integrating behavioral signals and counterfactual reasoning improves robustness within the defined protocol, the task formulation and evaluation remain constrained to dataset-specific labels. Despite their methodological differences, both systems face the same fundamental gap that strong performance on retrospective benchmarks does not establish readiness for prospective clinical use.

This limitation becomes more evident in clinical assessment settings that rely on structured, interactive evaluation rather than fixed-label classification. In psychiatric practice, many conditions are diagnosed through standardized clinician-administered interviews that capture longitudinal symptoms, contextual factors, and narrative responses. Anxiety and trauma-related disorders, such as post-traumatic stress disorder (PTSD), follow this paradigm and therefore introduce challenges that extend beyond those observed in depression modeling. Structured PTSD assessment relies on standardized clinician-administered interview protocols. TRUST approaches this as a dialogue management task aligned with the DSM-5 CAPS interview structure~\cite{tu2025trust}. The system implements 92 protocol variables through 241 predefined questions organized in rule-based dialogue trees. This structure ensures systematic coverage but limits adaptability when patient responses fall outside expected patterns. LLM integration does not fully resolve these constraints inherent to fixed-flow clinical dialogue systems. Human raters evaluating 100 sampled CAPS transcripts judged the generated agent dialogue to perform comparably to original interviews. In contrast, patient simulation fidelity received negative scores due to ongoing hallucination and information loss. The divergence between human and automated evaluation proved more concerning. LLM-based evaluators assigned substantially higher ratings than human experts, undermining confidence in claims of system readiness and highlighting the unreliability of fully automated assessment in this domain.

Related work addresses PTSD diagnosis through a different strategy based on data augmentation rather than interactive dialogue modeling. Wu et al. employ LLM-generated text to mitigate label imbalance in clinical interview transcripts, using both zero-shot generation of synthetic interviews and few-shot rephrasing of existing samples~\cite{wu2023automatic}. Within a controlled experimental protocol on the E-DAIC dataset, the augmented training sets yield consistent improvements in classification performance across multiple conventional classifiers. The primary contribution lies in demonstrating that large language models can generate interview-style text that is sufficiently coherent and label-consistent to improve in-dataset learning under severe class imbalance. However, the findings are based on retrospective analysis of a single dataset with a fixed evaluation split, and the study does not examine transfer across datasets, interview protocols, or clinical populations. As a result, it remains unclear whether the observed gains reflect improved coverage of clinically meaningful variation or merely increased alignment with dataset-specific linguistic patterns. The work establishes text augmentation as a practical tool for addressing class imbalance, but questions about cross-dataset generalization and the clinical validity of synthetic data remain unresolved.

\subsubsection{Neurodevelopmental disorders}
Neurodevelopmental disorders such as autism and ADHD may benefit from early detection and individualized interventions. AI models are being explored for potential applications in analyzing behavioral patterns and social interactions. Research is ongoing to develop adaptive support systems, though their clinical effectiveness requires further validation.

Across neurodevelopmental disorders, recent LLM-centric systems are increasingly framed as tools for structuring behavioral observations into screening or diagnostic support, although the strength of evidence varies substantially across tasks and study designs. Kulkarni et al. propose a video-based ADHD screening pipeline in which recorded behavioral sessions are decomposed into frame-level representations, followed by feature extraction and LLM-based contextual analysis of attention patterns and movement-related cues, before training a downstream classifier to produce high- or low-risk outputs~\cite{kulkarni2025utilizing}. The primary contribution is methodological, demonstrating how heterogeneous video-derived behavioral signals can be translated into structured representations by mapping low-level visual features into semantically organized behavioral descriptions via LLM-based analysis, which can then be consumed by downstream risk-stratification and visualization modules. In this sense, the work is most naturally interpreted as a study of representation and pipeline design for behavioral analysis, rather than as a performance-validated ADHD screening system.

In autism assessment, SocialRecNet targets a more narrowly defined construct by estimating ADOS social reciprocity scores from aligned speech and transcript segments using an LLM-guided multimodal architecture~\cite{chen2025socialrecnet}. The framework integrates turn-level acoustic features and textual content through cross-modal alignment and segment-level aggregation, allowing conversational dynamics to be summarized into representations that are compatible with LLM-based reasoning. Evaluation is reported on ADOS recordings from 105 children spanning typically developing controls and graded ASD severity strata, achieving a Pearson correlation of $r=0.711$ and an average improvement of 26.24\% over baseline methods. SocialRecNet shows how multimodal conversational signals can be organized to support continuous assessment of clinician-defined social reciprocity dimensions, rather than binary diagnosis. Despite these gains, generalization beyond the single-language ADOS setting remains uncertain, and the use of ADOS-derived scores as both supervision and evaluation anchors the model’s outputs to an existing clinical instrument.

Moving from assessment toward communication support, ASD-Chat introduces an LLM-driven dialogue intervention system whose prompt structure, topic sequencing, and turn-taking rules are derived from VB-MAPP~\cite{deng2024asd}. Rather than generating unconstrained conversational responses, the system embeds clinician-informed verbal behavior paradigms into the dialogue flow, using restricted question forms and staged topic progression to mirror common intervention practices. Empirical evaluation is conducted through short, topic-limited conversations with 12 children with ADOS-2 confirmed ASD, where dialogue text, speech characteristics, and fNIRS signals are analyzed alongside matched sessions led by professional interventionists. ASD-Chat demonstrates that structured LLM-mediated interaction can sustain engagement and elicit behavioral and physiological responses comparable to those observed in clinician-guided sessions. While the study shows how LLMs can implement established intervention protocols in resource-constrained settings, actual therapeutic impact has not been measured and requires prospective trials with clinically validated outcomes.

Across the disease domains discussed above, model performance appears closely linked to the structure of the evaluation protocol and the specificity of the clinical endpoint. Strong results are typically reported for well-bounded retrospective tasks, whereas evidence remains limited for longitudinal prognosis, cross-site robustness, and real-world workflow integration. The trajectory of disease-oriented AI may therefore hinge less on model scale alone than on the clarity of problem formulation and the rigor of prospective validation.

\section{Public datasets for computational neuroscience}
The availability of large and diverse public datasets from multiple neuroimaging modalities and clinical contexts has supported the growth of computational neuroscience research. These data encompass a range of techniques, including high-temporal-resolution electrophysiological recordings with millisecond precision and high-spatial-resolution structural and functional neuroimaging data relevant for anatomical and functional brain organization. Dataset scales vary from focused studies of dozens of participants to population-based studies with tens of thousands of subjects, collectively amounting to terabytes of neural data for statistical analyses and computational modeling. 

These datasets are used across a broad spectrum of neurological and psychiatric disorders, including neurodegenerative diseases, stroke, epilepsy, autism spectrum disorders, and mental health conditions, and they also support basic research in cognitive neuroscience, brain--computer interfaces, and neural decoding. Multimodal datasets that combine multiple recording techniques within the same subjects enable complementary analyses by leveraging the strengths of different modalities. Electrophysiological methods provide high temporal resolution, while neuroimaging methods offer spatial resolution. The availability of such multimodal datasets, together with standardized data formats and open-access longitudinal resources, has facilitated data sharing and reuse, supporting the development of computational approaches for studying brain function and dysfunction. We have compiled major datasets from the neuroscience literature over the past two decades, as summarized in Table~\ref{tab:neuroscience_datasets}. These public datasets vary widely in scope and scale and have supported progress in computational neuroscience by providing access to neural data across different recording modalities, subject populations, and experimental designs.

{\footnotesize
\setlength{\tabcolsep}{5pt}
\begin{longtable}{p{3.5cm} p{2.5cm} p{1.5cm} p{3.0cm} p{4.7cm}}
    \caption{Comprehensive overview of publicly available datasets for computational neuroscience research} \label{tab:neuroscience_datasets} \\
    \toprule
    \textbf{Dataset} & \textbf{Modality} & \textbf{Subjects} & \textbf{Data Size} & \textbf{Application Domain} \\
    \midrule
    \endfirsthead

    \caption[]{-- continued from previous page} \\
    \toprule
    \textbf{Dataset} & \textbf{Modality} & \textbf{Subjects} & \textbf{Data Size} & \textbf{Application Domain} \\
    \midrule
    \endhead

    \bottomrule
    \multicolumn{5}{r}{Continued on next page} \\
    \endfoot

    \bottomrule
    \endlastfoot

    \multicolumn{5}{c}{\textit{EEG Datasets}} \\
    \midrule
    TUH EEG Corpus~\cite{obeid2016temple}& EEG & 14,987+ & 1.7 TB / 21,787 h & General EEG analysis \\
    TUAB~\cite{lopez2015automated} & EEG & 2,383 & 1,139 h & Abnormality detection \\
    TUAR~\cite{hamid2020temple} & EEG & 213 & 83.7 h & Artifact detection \\
    TUSL~\cite{obeid2016temple} & EEG & 38 & 27.5 h & Slowing events classification \\
    TUEV~\cite{obeid2016temple} & EEG & 288 & -- & Clinical evaluation \\
    Siena Scalp EEG~\cite{detti2020siena} & EEG & 14 & 141 h & Scalp EEG analysis \\
    SEED-IV~\cite{zheng2018emotionmeter} & EEG & 15 & -- & Emotion recognition \\
    SEED-V~\cite{liu2021comparing} & EEG & 15 & 32.7 h & Emotion recognition \\
    DEAP~\cite{koelstra2011deap} & EEG & 32 & -- & Emotion recognition \\
    FACED~\cite{chen2023large} & EEG & 123 & -- & Affective computing \\
    CHB-MIT~\cite{goldberger2000physiobank} & EEG & -- & -- & Complex physiological signal analysis \\
    MIBCI~\cite{schalk2004bci2000} & EEG & 73 & -- & BCI research \\
    BCIC4-1~\cite{tangermann2012review} & EEG & 7 & -- & Motor imagery BCI \\
    Zuco~\cite{hollenstein2018zuco} & EEG & 12 & -- & BCI research \\
    Sleep-EDF~\cite{kemp2000analysis} & EEG & 22 & Two full-night/subject & Sleep analysis \\
    EEGMat~\cite{gifford2022large} & EEG & 10 & 82,160 trials & Material recognition \\
    STEW~\cite{44r8-ya50-18} & EEG & 48 & \textasciitilde4 h & EEG-based mental workload \\
    Go-Nogo~\cite{ds002680:1.0.0} & EEG & 14 & -- & Visual categorization \\
    MusicEEG~\cite{kaneshiro2016naturalistic} & EEG & 31 & -- & Temporal dynamics \\
    HBN EO/EC~\cite{alexander2017open} & EEG & 2,952 & -- & Resting state (pediatric) \\
    HBN-EEG~\cite{alexander2017open} & EEG & 1,897 & -- & Multiple tasks (pediatric) \\
    Features-EEG~\cite{ds004357:1.0.1} & EEG & 16 & -- & Visual feature processing \\
    HFO~\cite{cserpan2023dataset} & EEG & 30 & -- & Pediatric epilepsy \\
    PEARL-Neuro~\cite{dzianok2024pearl} & EEG & 79 & -- & Cognitive tasks \\
    RestCog~\cite{ds004148:1.0.1} & EEG & 60 & -- & Resting state cognition \\
    Awakening~\cite{ds005620:1.0.0} & EEG & 21 & -- & Sedation studies \\
    AD-Auditory~\cite{lahijanian2024auditory} & EEG & 33 & \textasciitilde5h & Auditory gamma entrainment in dementia \\
    ADFSU~\cite{vicchietti2023computational} & EEG & 184 & -- & Alzheimer's studies \\
    ADFTD~\cite{miltiadous2023dataset} & EEG & 88 & \textasciitilde19.4 h & Alzheimer \& FTD analysis\\
    ADSZ~\cite{alves2022eeg}~\cite{pineda2020quantile}& EEG & 109 & \textasciitilde36.3 h & Alzheimer's seizure analysis \\
    APAVA~\cite{escudero2006analysis}\cite{smith2017accounting} & EEG & 22 & \textasciitilde1.8h & Alzheimer’s EEG analysis \\
    BrainLat~\cite{prado2023brainlat} & EEG & 238 & \textasciitilde330 h & Latin American neurodegeneration \\
    Cognision-ERP~\cite{cecchi2015clinical} & EEG & 204 & \textasciitilde30 h & Alzheimer’s ERP biomarkers \\
    Cognision-rsEEG~\cite{ieracitano2019time} & EEG & 189 & \textasciitilde19 h & Alzheimer’s EEG classification \\
    CNBPM~\cite{amezquita2019novel} & EEG & 121 & \textasciitilde1.2 h & Alzheimer’s EEG analysis \\
    \midrule
    \multicolumn{5}{c}{\textit{MEG Datasets}} \\
    \midrule
    MEG-MASC~\cite{gwilliams2023introducing} & MEG & 27 & 2 h/subject & Natural speech MEG processing \\
    MEG-Narrative~\cite{armeni202210} & MEG & 3 & 10 h/subject & Narrative language MEG comprehension \\
    OMEGA~\cite{niso2016omega} & MEG & 644 & -- & Resting state + pathology \\
    CC700~\cite{taylor2017cambridge} & MEG & 700 & -- & Multiple cognitive tasks \\
    AversiveMEG~\cite{ds003682:1.0.0} & MEG & 28 & -- & Aversive learning \\
    MIND~\cite{ds004107:1.0.0} & MEG & 8 & -- & Somatosensory stimulation \\
    SMN4Lang~\cite{ds004078:1.2.1} & MEG & 12 & 6 h/subject & Language comprehension \\
    THINGS-MEG~\cite{ds004212:2.0.1} & MEG & 4 & -- & Object recognition \\
    ASWR-MEG~\cite{ds004276:1.0.0} & MEG & 24 & -- & Word sequence processing \\
    ImageLine~\cite{singer2023spatiotemporal} & MEG & 30 & -- & Visual object processing \\
    NeuroMorph~\cite{ds005241:1.1.0} & MEG & 24 & -- & Lexical decision tasks \\
    Kymata-SOTO~\cite{Thwaites_Zhang_Woolgar_Wingfield_Yang_2025} & MEG/EEG & 35 & -- & Multi-language conversations \\
    \midrule
    \multicolumn{5}{c}{\textit{MRI Datasets}} \\
    \midrule
    HCP~\cite{van2013wu} & MRI/MEG & 700+ & -- & Multi-modal brain imaging \\
    UK Biobank~\cite{collins2007uk} & MRI/fMRI & 61,038+ & 82,800 images & Population neuroimaging \\
    ADNI~\cite{jack2008alzheimer} & MRI/PET & 2,300+ & -- & Alzheimer's disease \\
    OASIS-2~\cite{marcus2010open} & MRI & 150 & 373 scans & Longitudinal brain aging \\
    OASIS-3~\cite{lamontagne2019oasis} & MRI/PET & 1,098 & 3,776 scans & Longitudinal brain aging \\
    BraTS-2015~\cite{menze2014multimodal} & MRI & 65 & 65 scans & Brain tumor segmentation \\
    BraTS-2019~\cite{bakas2018identifying} & MRI & 542 & 542 scans & Brain tumor segmentation \\
    BraTS-2021~\cite{baid2021rsna} & MRI & 2,040 & 8,000+ scans & Brain tumor segmentation \\
    ATLAS~\cite{liew2022large} & MRI & 1,271 & 1,271 scans & Stroke lesion analysis \\
    ABIDE I~\cite{di2014autism} & MRI & 1,112 & 1,112 scans & Autism spectrum disorder \\
    ABIDE II~\cite{di2017enhancing} & MRI & 1,044 & 1,044 scans & Autism spectrum disorder \\
    PPMI~\cite{marek2011parkinson} & MRI/DAT/CSF & 600 & -- & Parkinson's disease \\
    MCSA~\cite{roberts2008mayo} & MRI & 2,719 & 2,719 scans & MCI epidemiology \\
    SOOP~\cite{rorden2024stroke} & MRI &  2,888 &  2,888 scans & Brain development \\
    CBTN LGG~\cite{lilly2023children} & MRI & 4,700+ & -- & Pediatric brain tumors \\
    MIRIAD~\cite{malone2013miriad} & MRI & 69 & 708 scans & Dementia research \\
    DLBS~\cite{ds004856:1.0.0} & MRI & 464 & 957 scans & Brain aging \\
    UCSF-PDGM~\cite{calabrese2022university} & MRI & 495 & 124 GB & Glioma tumor analysis \\
    QIN-GBM~\cite{mamonov2016data} & MRI & 54 & 33.5 GB & Glioblastoma research \\
    UPENN-GBM~\cite{bakas2022university} & MRI & 630 & 357.42 GB & Glioblastoma research \\
    DFCI/BCH LGG~\cite{tak2024foundation} & MRI & 396 & 396 scans & Low-grade glioma \\
    DFCI/BCH HGG~\cite{tak2024foundation} & MRI & 534 & 534 scans & High-grade glioma \\
    RIDER~\cite{barboriak2015rider} & MRI & 19 & 7.79 GB & Glioblastoma research  \\
    wu1200~\cite{van2013wu} & MRI & 1,200 & 1,200 scans & Brain connectivity research \\
    LONG579~\cite{wang2022longitudinal} & MRI & 322 & -- & Longitudinal language development study \\
    BABY~\cite{nda2848} & MRI & 500 & -- & Infant brain connectome \\
    AOMIC~\cite{ds003097:1.2.1} & MRI & 1,370 & 5,738 scans & Multimodal MRI research \\
    Cananda-PD~\cite{xiao2017dataset} & MRI & 25 & -- & Canadian Parkinson’s disease \\
    HaN~\cite{podobnik2023han} & MRI/CT & 56 & 112 scans & Head \& neck organ-at-risk segmentation \\
    NIMH~\cite{hanson2011association} & MRI & 317 & 317 scans & Socioeconomic status and brain structure \\
    ICBM~\cite{mazziotta2001probabilistic} & MRI & \textasciitilde7,000 & \textasciitilde7,000 scans & Human brain atlas research \\
    IXI~\cite{ixi_dataset} & MRI & -- & \textasciitilde600 scans & MRI analysis in healthy populations \\
    PING~\cite{jernigan2016pediatric} & MRI & 1,493 & \textasciitilde5,000 scans & Pediatric imaging–genomics phenotyping \\
    Pixar~\cite{richardson2019mri} & MRI & 155 & 310 scans & Naturalistic social cognition \\
    SALD~\cite{wei2018structural} & MRI & 494 & \textasciitilde66 h & Brain aging research \\
    RadArt~\cite{nordstrom2018large} & MRI & 115 & -- & Radiation-induced cerebral arteriopathy \\
    Rhineland Study~\cite{breteler2014ic} & MRI & 30,000 & \textasciitilde25,000 h & Population brain aging study \\
    SchizConnect~\cite{wang2016schizconnect} & MRI & 1,129 & 21,309 volumes & Schizophrenia research \\
    OpenBHB~\cite{dufumier2022openbhb} & MRI & 5,330 & 5,330 scans & Brain age prediction debiasing \\
    MSSEG~\cite{commowick2018objective} & MRI & 42 & 212 volumes & Multiple sclerosis lesion segmentation \\
    Convers~\cite{rauchbauer2019brain} & MRI & 25 & -- &  Natural language neural decoding\\
    ISLES2022~\cite{hernandez2022isles} & MRI & 400 & 1,200 scans & Ischemic stroke lesion segmentation \\
    WMH2017~\cite{kuijf2019standardized} & MRI & 170 & 340 scans & White matter hyperintensities \\
    Multimodal MCI~\cite{yuan2025multimodal} & MRI & 40 & 40 & Cognitive mechanism analysis \\
    MSLesSeg~\cite{guarnera2025mslesseg} & MRI & 75 & 115 scans & Multiple Sclerosis Lesion Segmentation \\
    \midrule
    \multicolumn{5}{c}{\textit{Visual Datasets}} \\
    \midrule
    Natural Scenes~\cite{allen2022massive} & fMRI & 8 & -- & Visual-to-semantic mapping \\
    BOLD5000~\cite{chang2019bold5000} & fMRI & 4 & 15 scanning sessions & Natural image fMRI vision study \\
    GOD~\cite{horikawa2017generic} & fMRI & 5 & -- & Generic object decoding \\
    MS-COCO~\cite{lin2014microsoft} & Images & -- &  328,000 & Object recognition (stimuli) \\
    Flick30k~\cite{plummer2015flickr30k} & Images & -- & 31,783 & Image-caption pairs \\
    ConceptualCaptions12M~\cite{changpinyo2021conceptual} & Multi-modal & -- & 12,423,374 & Vision-language study \\
    ConceptualCaptions3M~\cite{sharma2018conceptual} & Multi-modal & -- & 3,369,218 & Vision-language study \\
    SBU~\cite{ordonez2011im2text} & Multi-modal & -- & \textasciitilde1,000,000 & Vision–language study \\
    VG~\cite{krishna2017visual} & Multi-modal & -- & 108,077 & Vision–language study \\
    Visual stimuli~\cite{rauchbauer2020multimodal} & fMRI & 594 & 53 time-steps/subject & Object recognition \\
    \midrule
    \multicolumn{5}{c}{\textit{Clinical and Disease-Specific Datasets}} \\
    \midrule
    PETfrog~\cite{ds002385:1.0.1} & PET/MR & 2 & 5 scans & PET–MRI neurodevelopment study \\
    ADReSSo~\cite{luz2021detecting} & Audio/Clinical & -- & 347 rercordings & Alzheimer's speech analysis \\
    NCCT~\cite{umerenkov2025core} & CT & 112 & 112 scans & Hyperacute stroke segmentation \\
    Dreaddit~\cite{turcan2019dreaddit} & Text & -- & 187,444 posts & Depression detection \\
    CSSRS-Suicide~\cite{gaur2019knowledge} & Text & 2,181 users & 15,755 posts & Suicide risk assessment \\
    PPAT~\cite{wu2024vs} & Behavioral clinical & 690 & 690 & Depression assessment \\
    DepSeverity~\cite{naseem2022early} & Text & 3,743 users & 80,903 posts & Depression severity classification\\
    Semio2Brain~\cite{alim2022probabilistic} & Clinical & 4643 & 11,230 samples & Seizure semiology\\
    SDCNL~\cite{haque2021deep} & Text & -- & 1,895 posts & Suicide \& depression analysis \\
    PMData~\cite{thambawita2020pmdata} & Clinical & 16 & 5 months of logging & Mental health monitoring \\
    Globem~\cite{xu2022globem} & Behavioral clinical & 497 & -- & Passive-sensing depression detection \\
    E-DAIC~\cite{devault2014simsensei} & Audio/Clinical & 351 & 351 sessions & Depression interview \\
    SeeKr~\cite{yang2026knowledge} & Clinical & -- & 852 samples & Seizure semiology \\
    % ADOS~\cite{lord2000autism} & Clinical & 381 & -- & Autism diagnosis \\
    CQ500 ~\cite{chilamkurthy2018deep} & CT& 491 & 491 scans & head trauma or stroke analysis \\
    Psych-101~\cite{binz2024centaur} & Psychological & 60,092 & 10,681,650 samples & Computational psychiatry \\
    ClinicalQA~\cite{zakka2024almanac} & Clinical & 5 reviewers & 130 Q\&As & Clinical decision support research\\
    \midrule
    \multicolumn{5}{c}{\textit{Specialized Research Datasets}} \\
    \midrule
    DANDI~\cite{perich2025long} & Neuronal spikes & 2 primates & \textasciitilde78 sessions & Motor function analysis \\
    \midrule
    \multicolumn{5}{c}{\textit{Knowledge, Literature Resources and Other Databases}} \\
    \midrule
    PubMed & Literature & -- & -- & Neuroscience literature \\
    Wikipedia & Encyclopedia & -- & -- & General knowledge \\
    NeuroLex~\cite{larson2013neurolex} & Knowledge & -- & -- & Neuroscience ontology \\
    NeuroMorpho~\cite{ascoli2007neuromorpho} & Morphology & -- & -- & Neuronal morphology \\
    OpenNeuro~\cite{markiewicz2021openneuro} & Multi-modal neuroimaging data & -- & -- & Comprehensive neuroimaging research\\
    Neurosurgical Atlas~\cite{teton2020neurosurgical} & Literature & -- & -- & Neurosurgical atlas \\
    NLI4CT~\cite{jullien2023nli4ct} & Clinical text & -- & -- & Clinical trial inference \\
    Merck Manual~\cite{merckmanual} & Medical & -- & -- & Medical reference \\
    Neurology textbooks & Literature & -- & -- & Educational resources
\end{longtable}
}

\section{Future development and challenges}

Section~\ref{sec:applications} highlights rapid progress across neuroimaging, electrophysiology, and language-guided neuroreasoning systems. The main barrier to impact is no longer whether large-scale models can fit benchmark datasets, but whether they can be trained, evaluated, and deployed under realistic heterogeneity, privacy constraints, and clinical workflow requirements. As presented in Fig.~\ref{fig:eval_gap}, this gap persists across modalities. The horizontal axis represents model performance as reported in original benchmark evaluations, whereas the vertical axis reflects deployment readiness, incorporating validation breadth, dataset diversity, and the extent of cross-dataset or cross-site evaluation. In this context, model positions reflect the scope and strength of publicly reported validation evidence rather than direct numerical comparisons across heterogeneous studies, tasks, or modalities, and should therefore be understood as indicative of relative developmental maturity rather than as a ranking of overall model quality or clinical usefulness. Although Table~\ref{tab:neuroscience_models} indicates that researchers are increasingly leveraging multi-center data to improve robustness, the universal absence of prospective validation highlights that the field has not yet crossed the threshold from retrospective benchmarking to real-world clinical deployment. This section therefore shifts from describing challenges to outlining technical and governance actions that can be implemented, scrutinized, and sustained in practice. To provide a structured path forward, we synthesize the challenges discussed in the following subsections into two distinct horizons: (1) Short-term Engineering and Validation Priorities, which address immediate hurdles in robustness and efficiency; and (2) Long-term Methodological Frontiers, which represent fundamental needs for mechanistic insight and closed-loop adaptation. Table~\ref{tab:future_roadmap} summarizes this strategic roadmap.

\subsection{Multimodal alignment and integration}

Multimodal FMs promise unified representations across imaging, signals, and text, yet biomedical multimodal data are typically fragmented, inconsistently preprocessed, and unevenly distributed across sites and protocols~\cite{liu2025challenges}. As a result, progress in multimodal alignment is best assessed against failure modes that routinely arise in neuroscience, including partial modality availability, protocol drift, and site-specific acquisition artifacts.

In practice, effective multimodal systems tend to emphasize robustness to incompleteness rather than assuming comprehensive input availability.
\begin{itemize}
    \item \textbf{Design for missing modalities.} Training and evaluation should explicitly report performance under structured missingness patterns, such as single-modality, paired-modality, and clinically realistic subsets, rather than relying on idealized complete inputs. Architectures that support conditional inference when only a subset of inputs is available are therefore preferable.
    \item \textbf{Harmonization as an explicit modeling objective.} When data are pooled across sites or protocols, harmonization should be treated as a first-class component and evaluated for its ability to reduce site-specific leakage while preserving task-relevant signal.
\end{itemize}
Beyond these considerations, alignment objectives benefit from being grounded in neuroscientific structure. Supervision or self-supervision tied to anatomical correspondence, functional network organization, or clinically defined endpoints can introduce inductive constraints that improve robustness, provided their contributions are isolated through targeted ablation analyses.

\subsection{Computational efficiency and scalability}

Model scale alone is not a sufficient indicator of usability in research or clinical settings. Practical constraints arise from the cumulative cost of preprocessing, training, inference latency, and maintenance. Approaches that prioritize compact architectures, modular encoders, and efficient tokenization are therefore more likely to support reproducible and transferable use~\cite{zhang2023brant,luo2024large}.

From an engineering standpoint, several considerations repeatedly emerge.
\begin{itemize}
    \item \textbf{Parameter-efficient adaptation.} Adapter-based or low-rank fine-tuning strategies should be reported alongside full fine-tuning, with explicit accounting of memory usage and latency on representative hardware.
    \item \textbf{Distillation for routine deployment.} Large models can serve exploratory or adjudication roles, while task-specific distilled variants support routine inference with lower computational overhead.
    \item \textbf{Compute-aware evaluation.} Accuracy gains are most informative when reported together with inference time, memory footprint, and energy proxies, enabling meaningful comparison across deployment contexts.
\end{itemize}

{\footnotesize
\setlength{\tabcolsep}{7pt}
\begin{longtable}{p{3.3cm} p{5.2cm} p{6.5cm}}
    \caption{Roadmap of key challenges and strategic solutions for AI in neuroscience} \label{tab:future_roadmap} \\
    \toprule
    \textbf{Challenge Domain} & \textbf{Current Bottleneck} & \textbf{Potential Strategic Solutions} \\
    \midrule
    \endfirsthead

    \caption[]{-- continued from previous page} \\
    \toprule
    \textbf{Challenge Domain} & \textbf{Current Bottleneck} & \textbf{Proposed Path Forward} \\
    \midrule
    \endhead

    \bottomrule
    \multicolumn{3}{r}{Continued on next page} \\
    \endfoot

    \bottomrule
    \endlastfoot

    % === Horizon 1 ===
    \multicolumn{3}{c}{\textbf{\textit{Short-term Engineering \& Validation Priorities}}} \\
    \midrule
    \textbf{Robust Evaluation} & Overestimation of performance due to random splits; lack of prospective evidence. & Adoption of \textbf{deployment-mimicking splits} (site/time-held-out) and transition to pre-registered \textbf{prospective clinical trials} (Section 5.4, 5.6). \\
    \textbf{Data \& Privacy} & Fragmented inputs and privacy risks in centralized training. & Design for \textbf{missing modalities} and implementation of \textbf{Federated Learning} with explicit privacy-utility reporting (Section 5.1, 5.7). \\
    \textbf{Efficiency} & High latency and memory footprint limiting clinical workflow integration. & Shift to \textbf{task-specific distilled variants} and parameter-efficient adaptation (e.g., LoRA) for routine deployment (Section 5.2). \\
    \textbf{Governance} & Algorithmic bias and undetected model errors in deployment. & Rigorous \textbf{bias assessment} for underrepresented populations and mandatory \textbf{post-deployment monitoring} for drift (Section 5.7). \\
    \midrule
    
    % === Horizon 2 ===
    % === Horizon 2 ===
    \multicolumn{3}{c}{\textbf{\textit{Long-term Fundamental Methodological Frontiers}}} \\
    \midrule
    \textbf{Mechanistic Insight} & Models function as black boxes; post-hoc explanations are unreliable. & Validation against falsifiable \textbf{mechanistic targets} (e.g., network motifs, lesion-consistency) rather than just accuracy (Section 5.3). \\
    \textbf{Advanced Alignment} & Integration objectives often ignore biological structure. & Incorporation of inductive constraints tied to \textbf{anatomical correspondence} and functional organization into supervision (Section 5.1). \\
    \textbf{Human-AI Collaboration} & Systems lack operational boundaries or "common sense." & Development of workflows with explicit \textbf{operational constraints} and evidence-oriented interfaces for verifiability (Section 5.5). \\
    \textbf{Digital Twin Brain} & Predominantly static AI models lack longitudinal updating and intervention-aware simulation. & Development of \textbf{individualized computational brain models} that integrate multi-scale biological and imaging data, support dynamic updating, and enable in \textbf{model-based simulation of intervention effects} (Section 5.8).\\
    \textbf{Emerging Frontiers} & Static models fail in dynamic, real-time neural environments. & Evolution toward \textbf{closed-loop neurotechnology} capable of safe online adaptation and generative simulation (Section 5.9). \\

\end{longtable}
}

\subsection{Interpretability, uncertainty, and mechanistic insight}

Interpretability functions as a requirement for translation rather than a stylistic preference. Clinical users must be able to assess why a model supports a conclusion, what evidence may be missing, and when the system operates outside its competence. Prior analyses emphasize that explanations should enable interrogation of decisions, particularly in safety-critical settings where black-box performance alone is insufficient~\cite{dinsdale2022challenges,rudin2019stop}. While anatomical or mechanistic constraints can reduce degrees of freedom, they do not substitute for validation of explanatory claims~\cite{barbano2024anatomical,wang2025lead}.

Interpretability becomes actionable when it is evaluated rather than asserted.
\begin{itemize}
    \item \textbf{Distinguishing decision-relevant evidence from post-hoc justification.} Feature attribution or saliency methods should be accompanied by stability checks and sanity tests, and complemented with example-based retrieval or counterfactual analysis when feasible~\cite{miller2019explanation}.
    \item \textbf{Uncertainty as an output.} Calibration and abstention behavior should be reported explicitly, with predefined mechanisms for escalation to human review, additional testing, or conservative fallback strategies.
\end{itemize}
When models are claimed to provide neuroscientific insight, evaluation should include falsifiable mechanistic targets, such as reproducible network motifs, known stimulus--response relationships, or lesion-consistent degradations, assessed separately from task accuracy. 

An additional dimension that remains underexplored in current neuro-AI systems is the integration of genetic and molecular information. Most existing large-scale models in neuroscience are trained on imaging, electrophysiological, or behavioral data, which primarily capture system-level phenotypes. While these representations are highly informative for prediction, their ability to support mechanistic interpretation and individualized modeling is inherently limited without links to underlying molecular processes. Genetic variation, cell-type–specific transcriptomic profiles, and molecular pathway information provide a complementary axis of biological organization that can constrain model representations, improve cross-subject interpretability, and enable stratification beyond phenotype-level similarity. From a foundation-model perspective, such data can serve as structured priors for multimodal alignment, as anchors for biologically meaningful latent spaces, or as targets for cross-scale representation learning. However, the incorporation of these modalities is currently restricted by limited cross-scale datasets and unresolved questions regarding harmonization and privacy. In this sense, genetic and molecular integration is less an additional application domain than a key methodological pathway for grounding model representations in biologically interpretable mechanisms, thereby enabling falsifiable and cross-subject-consistent neuroscientific insight.

\subsection{Generalization, robust evaluation, and dataset shift}

A persistent source of failure in deployment is dataset shift induced by scanner differences, protocol variation, demographic composition, and temporal changes in clinical practice~\cite{leming2023challenges,dinsdale2022challenges}. Because many neuroscience benchmarks are internally harmonized, random train--test splits can substantially overestimate real-world robustness.

Empirical studies have documented substantial performance degradation when models trained on homogeneous research cohorts are applied to heterogeneous real-world clinical data~\cite{leming2023challenges}. Furthermore, "shortcut learning" remains a pervasive failure mode; Dinsdale et al.~\cite{dinsdale2022challenges} demonstrated that deep learning models may inadvertently exploit scanner-related confounders rather than biological signals, leading to degraded generalization across sites and highlighting the need for explicit harmonization strategies. Instances of negative transfer have also been observed, particularly where FMs pre-trained on natural images introduce texture-based biases that are suboptimal for tasks requiring precise volumetric quantification.

Evaluation strategies that approximate deployment conditions therefore provide more informative signals.
\begin{itemize}
    \item \textbf{Deployment-mimicking splits.} Site-held-out, scanner-held-out, and time-held-out evaluations offer a closer approximation to clinical use, particularly when complemented by subgroup analyses across relevant covariates.
    \item \textbf{Stress testing under plausible perturbations.} Controlled perturbations and test-time augmentation can be used to probe sensitivity to acquisition variability and to identify failure thresholds rather than average-case behavior.
    \item \textbf{Transparent adaptation and harmonization.} When domain adaptation or personalized federated learning is employed, studies should describe what information is shared, how privacy is preserved, and how performance varies across sites~\cite{huang2025ai}.
\end{itemize}

\subsection{Human--AI collaboration and cognitive modeling}

Language- and knowledge-guided systems increasingly support neuroscientists and clinicians through structured querying, atlas-grounded retrieval, and interactive interpretation workflows~\cite{hopkins2024atlasgpt,guo2023neurogpt}. The central question is not whether such systems can generate plausible responses, but whether they improve decision quality, reduce time-to-answer, and make errors easier to detect under realistic workloads.

Effective collaboration typically depends on clear operational boundaries.
\begin{itemize}
    \item \textbf{Decision support with explicit operational constraints.} System roles should be explicitly defined, including what recommendations are permitted, what actions are excluded, and which situations require mandatory human confirmation.
    \item \textbf{Human-centered evaluation.} Beyond model-centric metrics, evaluation should consider time savings, inter-rater agreement, and error detectability, ideally through prospective or workflow-integrated studies~\cite{huang2025ai}.
    \item \textbf{Evidence-oriented interfaces.} Outputs that surface citations, uncertainty, and provenance allow users to verify claims and reduce the risk of silent failure.
\end{itemize}

\subsection{Regulation, standardization, and benchmarking}

Translation also depends on shared standards for reporting, validation, and model updating. Clinical diagnostic AI faces technical, disease-specific, and institutional barriers, and progress is often most plausible through local deployment that demonstrably improves workflow, combined with close collaboration between clinicians and data scientists~\cite{leming2023challenges}. For adaptive systems, governance must account for monitoring, drift response, and change management rather than assuming static approval~\cite{dinsdale2022challenges}.

At a community level, several practices consistently recur.
\begin{itemize}
    \item \textbf{Benchmarking beyond accuracy.} Calibration, subgroup performance, and failure-mode characterization provide a more complete picture of system behavior than aggregate accuracy alone~\cite{cox2024brainsegfounder,tak2024foundation}.
    \item \textbf{External and prospective validation.} Retrospective internal benchmarks are best treated as feasibility checks, with multi-site external validation serving as a minimum bar for translational claims~\cite{leming2023challenges}.
    \item \textbf{Post-deployment monitoring.} Drift detection, periodic audits, and clearly assigned responsibility for intervention are necessary components of responsible deployment.
\end{itemize}

\subsection{Data governance, privacy, and responsible deployment}

Neuroscience datasets contain high-dimensional personal information and remain vulnerable to linkage attacks and re-identification despite standard anonymization procedures~\cite{dinsdale2022challenges}. Recent reviews emphasize the coupled risks of algorithmic bias and privacy leakage, particularly when models are trained or evaluated across institutions~\cite{thomas2024assessing}.

Responsible deployment therefore requires governance mechanisms that extend beyond technical performance.
\begin{itemize}
    \item \textbf{Representativeness and bias assessment.} Risk analyses should identify underrepresented populations and motivate targeted data collection, reweighting strategies, or carefully validated synthetic augmentation when appropriate~\cite{sun2025generating}.
    \item \textbf{Privacy-preserving training with disclosure.} Federated learning and differential privacy offer viable alternatives when data cannot be centralized, but their privacy--utility trade-offs should be reported explicitly, particularly for rare or clinically important patterns~\cite{huang2025ai}.
\end{itemize}
Clear documentation practices, including model cards, intended-use statements, and versioned data descriptions, are essential for communicating limitations and ensuring accountability across updates.

\subsection{Prospects for a digital twin brain}
\label{sec:digital_twin_brain}

The concept of the digital twin brain has recently been articulated as a framework for individualized, data-informed brain modeling~\cite{xiong2023digital}. Originating from systems engineering, a digital twin denotes a computational counterpart of a physical system that is continuously updated using real-world data and capable of simulation under hypothetical scenarios. In neuroscience, this idea has been proposed as a bridge between biological and artificial intelligence, aiming to construct subject-specific models that integrate structural and functional information for prediction and intervention analysis. More recent discussions emphasize that a digital twin brain is not a single algorithm but a systems-level construct composed of three elements: a personalized structural scaffold, a generative dynamical model capable of reproducing observable signals, and an inference mechanism that allows parameters and latent states to be updated as new data are acquired~\cite{wang2024virtual}. This framing highlights the central methodological challenge of solving high-dimensional inverse problems while preserving physiological plausibility and clinical interpretability.

Within this architecture, artificial intelligence methods are often positioned as enabling technologies for inference and model calibration. In particular, recent analyses argue that machine learning can assist inverse modeling, parameter estimation, and data assimilation in complex neural systems~\cite{nogaret2025ai}. Foundation models may therefore contribute to digital twin pipelines by providing robust feature extraction from heterogeneous modalities or amortized inference for high-dimensional parameter spaces. However, these contributions address specific computational components rather than constituting a complete digital twin system. A complementary line of work explores how digital twin modeling may support regenerative strategies and controlled modulation of neural dynamics, underscoring the importance of coupling data-driven approaches with mechanistic constraints~\cite{tataru2025designing}. This perspective reinforces the view that predictive accuracy alone is insufficient; stability, interpretability, and controllability remain essential for translational use.

At the same time, recent discussions indicate that digital twins in neuroscience remain primarily conceptual and have yet to undergo rigorous clinical validation~\cite{sandrone2024digital}. Longitudinal updating, cross-site robustness, and prospective evaluation protocols are still underdeveloped in most existing AI-driven neuroscience systems. In this sense, while FMs may strengthen representation learning and inference within future twin architectures, the realization of operational digital twin brain systems will depend as much on validation design and mechanistic grounding as on model scale.

\subsection{Emerging frontiers}

Complementary to the systemic vision of the digital twin brain, several other directions are likely to shape the next phase of neuroscience AI. Closed-loop and on-device neurotechnology places increasing emphasis on low-latency inference, uncertainty-aware control, and safe online adaptation~\cite{stanojevic2024high}. Generative modeling offers opportunities for data augmentation, counterfactual testing, and privacy-aware simulation, but requires clear separation between synthetic evaluation tools and clinical evidence~\cite{sun2025generating}. Sustained progress will also depend on cross-disciplinary standards and training that bridge neuroscience, clinical practice, and machine learning engineering~\cite{eke2022international,luppi2024trainees}.

\section{Conclusion}

The development of large-scale AI models in neuroscience is gradually expanding our ability to characterize brain function and support clinical decision-making. This survey has reviewed the diverse range of such models, from neural signal processing architectures used for decoding neural activity to clinical decision support systems that incorporate multimodal biomedical information. More fundamentally, this evolution reflects a shift from task-specific architectures toward generalist, representation-centered systems that can be adapted across modalities, datasets, and clinical contexts. These developments indicate that large-scale AI models are increasingly used not only as analytical tools but also as practical components in research workflows and early-stage translational settings. The progress summarized here reflects a gradual shift from isolated proof-of-concept demonstrations toward more systematic applications with measurable relevance to both foundational and applied neuroscience.

At the same time, fundamental challenges remain unresolved. Technical barriers include limited robustness to distribution shifts across heterogeneous clinical populations and recording conditions, difficulties in aligning multimodal data with heterogeneous temporal resolutions and preprocessing pipelines, and persistent gaps between neural patterns learned by models and the underlying cognitive or pathological mechanisms they aim to capture. Translational deployment faces additional obstacles: prospective clinical validation through pre-registered trials remains largely absent, and retrospective benchmark performance has not been shown to predict real-world outcomes. Ethical considerations, including patient privacy, algorithmic bias, and the appropriate use of AI-generated recommendations, require interdisciplinary frameworks that balance innovation with safety. Progress will require sustained collaboration across neuroscience, computer science, clinical medicine, and regulatory policy. While large-scale AI models hold meaningful potential to support both research and clinical practice, their long-term impact will depend on rigorous prospective validation, transparent dataset governance, and careful translational oversight that recognizes the gap between controlled experimental success and safe, effective clinical deployment.

\section*{Acknowledgments} The authors are grateful to Dr. Alison Anderson and  Mr. Dinesh  Giritharan from Monash University for their helpful discussion in genomics foundation models. 

\section*{Funding declaration}
This effort was partially supported by the National Institute of Neurological Disorders and Stroke of the National Institutes of Health, United States under Award Number R21NS135482 (PI: Liu). The content is solely the responsibility of the authors and does not necessarily represent the official views of the National Institutes of Health.

\section*{Competing interests}
The authors declare no competing interests.

\section*{Data availability}
All data analyzed in this study are derived from previously published studies and are available through the original sources cited in the reference list.

\section*{Ethics statement}

This study is a review article and does not involve human participants, animal experiments, or the use of newly collected clinical data. All information is derived from previously published studies and publicly available resources. Ethical approval and informed consent were therefore not required.

\bibliographystyle{ieeetr}
\bibliography{sample}

\end{document}